\pdfoutput=1
\documentclass[11pt]{article}

\usepackage{acl}

\usepackage{times}
\usepackage{latexsym}
\usepackage{amsmath} 
\usepackage{amssymb}
\usepackage{amsfonts}

\usepackage{booktabs}

\usepackage{xcolor}
\usepackage{colortbl}
\usepackage{hyperref}

\usepackage[T1]{fontenc}
\usepackage[utf8]{inputenc}

\usepackage{microtype}

\usepackage{inconsolata}

\usepackage{graphicx}

\usepackage[most]{tcolorbox}

\title{Curiosity Across Cultures: Evaluating Human–LLM Alignment with CUEST}
\title{The Curious Case of Curiosity across Human Cultures and LLMs}

\author{Angana Borah$^1$\hspace{5pt} 
Zhijing Jin$^{2, 3,4}$\hspace{5pt}  
 Rada Mihalcea$^1$  \\
$^1$University of Michigan - Ann Arbor, USA  \\
$^2$University of Toronto
$^3$Vector Institute \\
$^4$MPI for Intelligent Systems, Tubingen, Germany \\
\textit{\{anganab, mihalcea\}@umich.edu} \\  }
\usepackage{longtable}

\begin{document}
\maketitle
\begin{abstract}
Recent advances in Large Language Models (LLMs) have expanded their role in human interaction, yet curiosity -- a central driver of inquiry -- remains underexplored in these systems, particularly across cultural contexts. In this work, we investigate cultural variation in curiosity using Yahoo! Answers, a real-world multi-country dataset spanning diverse topics. We introduce \textit{CUEST} \textit{(\textbf{CU}riosity \textbf{E}valuation across \textbf{S}ocie\textbf{T}ies)}, an evaluation framework that measures human-model alignment in curiosity through linguistic (style), topic preference (content) analysis and grounding insights in social science constructs. Across open- and closed-source models, we find that LLMs flatten cross-cultural diversity, aligning more closely with how curiosity is expressed in Western countries. We then explore fine-tuning strategies to induce curiosity in LLMs, narrowing the human-model alignment gap by up to 50\%. Finally, we demonstrate the practical value of curiosity for LLM adaptability across cultures, showing its importance for future NLP research.

\end{abstract}

\section{Introduction}

Curiosity is central to human learning, driving exploration, and questioning that often yield to innovation and cultural insights~\cite{gruber2014states}. Across several disciplines, such as social psychology~\cite{berlyne1954theory}, and cognitive science~\cite{hebb1955drives}, curiosity is considered an internal drive that propels us toward new knowledge and experiences. Recent studies emphasize curiosity’s role in cultural learning~\cite{mikhaylov2016curiosity}, education~\cite{lindholm2018promoting, markey2014curiosity}, and human–technology interaction~\cite{ganuthula2024curiosity}. Studies also show that curiosity varies across societies, influenced by social norms and educational contexts~\cite{bluemke2024measuring, jirout2024curiosity}, thus framing it as both a cognitive–emotional driver and a socio-cultural process shaping human interaction. 
 
% Several theories of curiosity, for example, the Information Gap Theory~\cite{loewenstein1994psychology}, Curiosity Drive Theory~\cite{berlyne1960conflict}, and Optimal Arousal Theory~\cite{hebb1955drives}, emphasize different mechanisms through which curiosity operates: from reducing gaps in knowledge, to internal drives, and maintaining an optimal level of stimulation. More recent studies focus on curiosity’s role in cultural learning~\cite{mikhaylov2016curiosity}, education~\cite{lindholm2018promoting, markey2014curiosity}, and human–technology interaction~\cite{ganuthula2024curiosity}, framing it as not only a cognitive-emotional driver but also a socio-cultural process that shapes how people engage with each other.
\definecolor{pastellavender}{RGB}{230, 230, 250} % light lavender
\definecolor{pastelteal}{RGB}{200, 240, 240}     % soft teal
\definecolor{pastelpeach}{RGB}{255, 229, 180}    % warm peach
\definecolor{pastelmint}{RGB}{220, 255, 220}     % mint green
\definecolor{pastelsky}{RGB}{210, 235, 255}      % sky blue
\definecolor{pastelrose}{RGB}{255, 210, 220}     % soft rose pink
\definecolor{pastelyellow}{RGB}{255, 250, 205}   % light yellow
\definecolor{pastelgray}{RGB}{240, 240, 240}     % neutral gray

\begin{figure}
\centering
\includegraphics[width=\linewidth]{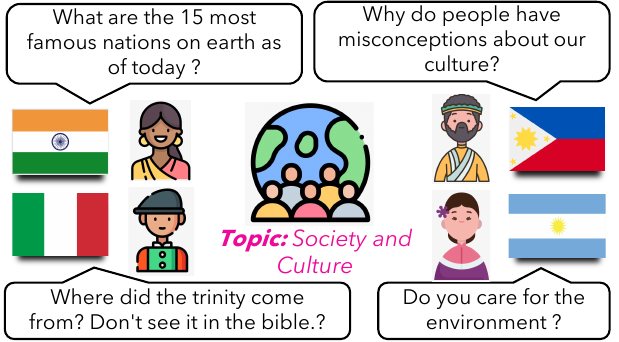}
\caption{\textbf{Curiosity-driven question differences across countries.} For a given topic, curiosity varies in style and content across countries. Here, we show examples from Argentina, India, Italy, and the Philippines. We extend this analysis to 18 countries across 16 topics.}
\vspace{-20pt}
\label{fig:first_diag}
\end{figure}

Moreover, people interact with LLMs primarily through dialogue, where the model's capacity to ask questions or simulate curiosity could effectively shape user experience and utility~\cite{ganuthula2024curiosity}. Yet, while curiosity is deeply influenced by cultural norms and communication styles~\cite{mikhaylov2016curiosity}, most cross-cultural studies test LLM knowledge by focusing on answering capabilities~\cite{seth-etal-2024-dosa, li2024culturellm} using survey-based benchmarks (e.g., WVS,\footnote{https://www.worldvaluessurvey.org/wvs.jsp} Pew Research.\footnote{https://www.pewresearch.org/}) Only limited prior work has examined curiosity through LLM's question-asking capability, but they are largely unconstrained by cultural nuance~\cite{ceraolo2024analyzing}. Guided by these limitations, we pose three research questions: (1) Do curiosity-driven questions on online platforms vary across cultures in humans, and do LLMs reproduce these patterns?, (2) What are some approaches to induce curiosity in LLMs? and (3) What are the practical implications of culture-aware curiosity for downstream applications of LLMs? 

Unlike prior work that examines cultural biases in LLMs in isolation, we believe ours is the first to directly compare human and LLM \textit{questions} across cultures using a systematic evaluation framework. The key contributions of our work are: (1) We introduce \textbf{CUEST}: a framework combining linguistic and content analysis with cultural grounding to analyze human and LLM curiosity questions across cultures. We also assess whether curiosities reflect traditional/stereotypical cultural tendencies as conceptualized in social science theories. (2) We explore various fine-tuning strategies to induce curiosity in LLMs, and evaluate the culture-aware curiosity and inherent question-asking abilities of LLMs, and (3) We examine the practical implications of curiosity on several cross-cultural datasets, demonstrating that it enhances LLMs’ adaptability in cultural norm understanding. Our contributions show the importance of curiosity in enhancing cross-cultural LLM capabilities. 
% footnote{While curiosity can manifest in many forms, we use question generation in this work.}
% \begin{enumerate}
% \item 
% \item 
% \end{enumerate}

\section{Related Work}
% \subsection{}
Psychological research on curiosity has long recognized its role as a key driver of exploration and learning. Several theories: Information Gap Theory~\cite{loewenstein1994psychology}, Curiosity Drive Theory~\cite{berlyne1960conflict}, and Optimal Arousal Theory~\cite{hebb1955drives}, emphasize different mechanisms through which curiosity operates: from reducing gaps in knowledge, to internal drives for uncertainty resolution, and maintaining an optimal level of stimulation. Studies also show that curiosity is important for cultural learning~\cite{mikhaylov2016curiosity}, education~\cite{lindholm2018promoting, markey2014curiosity}, and human–technology interaction~\cite{ganuthula2024curiosity}, therefore making a socio-cultural driver that shapes how people engage with each other. Finally, studies show that curiosity levels and expressions vary across societies, influenced by social norms, cultural values, and educational contexts~\cite{bluemke2024measuring, jirout2024curiosity}. 

In NLP, curiosity is only recently gaining attention.~\citet{ceraolo2024analyzing} propose the NatQuest corpus, showing that natural questions are often more open-ended and causally oriented than benchmark tasks. However, cultural variation in curiosity remains unexplored. Conversely, Prior work on cultural representation in LLMs highlight biases and call for culture-aware evaluation~\cite{seth-etal-2024-dosa, li2024culturellm, arora2022probing, borah-etal-2025-towards}, but do not examine models’ question-asking abilities. Our framework is the first to systematically compare cross-cultural human-LLM questions and propose methods to induce curiosity and improve cultural adaptability in LLMs.

\section{Cross-Cultural Questions Dataset}

% - brief intro paragraph to state we use cross-cultural human data and LLM data
% Curiosity is a multifaceted construct, yet it remains unified by its central role in driving exploration and knowledge seeking.

To evaluate culture-aware curiosity, we use two complementary sources: (1) human-authored questions and preferences, and (2) LLM-generated questions and topic preferences (using cultural personas). Together, these datasets allow for a systematic comparison of cultural differences in human and LLM curiosity. Table~\ref{tab:examples_questions} shows some examples.
% from the cross-cultural questions dataset. 

\begin{table}[h]
\centering
\small
\begin{tabular}{@{}lp{2.5cm}p{3cm}@{}}
\toprule
\textsc{Country} & \textsc{Human} & \textsc{LLM (LLaMA-3-8b)} \\
\midrule
\multicolumn{3}{c}{\textit{Topic: Family \& Relationships}} \\
\midrule
Argentina & Do I miss someone with whom I didn't have a relationship? & How to balance individual freedom with family responsibilities? \\
\addlinespace
Canada & Why are children these days not playing together outdoors? & How to navigate the complexities of blended families and step-parenting? \\
\midrule
\multicolumn{3}{c}{\textit{Topic: Society \& Culture}} \\
\midrule
France & Does the multiplication of the arts give a culture nectar? & What role do public holidays and celebrations play in French society? \\
\addlinespace
Singapore & Which is more dangerous? Corona virus or the effect of climate change? & Can you discuss the impact of globalization on our culture and preservation of local traditions? \\
\bottomrule
\end{tabular}
\caption{\textbf{Cross-Cultural Questions Dataset:} Question examples for two topics across four countries.}
\label{tab:examples_questions}
\vspace{-20pt}
\end{table}

\subsection{Human-authored Questions}
We use the ``Yahoo! Answers'' dataset,\footnote{https://www.quantcdn.io/blog/articles/we-archived-yahoo-answers} which contains user-submitted questions and answers from several countries across Asia, Europe, the Americas, and Australia. It spans diverse topics: Arts \& Humanities, Health, Education, etc., reflecting varied cultural perspectives. For our study, we include only those topics that have questions from all countries. Questions in local languages (e.g., Spanish in Argentina, etc.) were translated into English using the Google Translate API,\footnote{\url{https://cloud.google.com/translate}} enabling comparison while preserving meaning and structure. Additionally, we derive topic preferences by the frequency of questions per topic for each country, treating higher frequencies as indicators of stronger national interest. Overall, our study spans 16 topics across 18 countries (see Appendix~\ref{sec:yahoo} for details). 

\subsection{LLM-generated Questions}
\label{cur_eval}

We perform the following two tasks to generate LLM-based questions and preferences: \textbf{1) LLM-Question Generation}: We instruct LLMs to adopt a country persona and employ the following prompt structure for a given topic in the Yahoo dataset: \texttt{``Assume you are from <country>. Based on your cultural context in <country>, generate <x> questions related to <topic>.''}. This task aims to capture the stylistic forms of curiosity. \textbf{2) Topic Preference Analysis}: We prompt country-based LLM personas to generate topic rankings: \texttt{``You are representing the cultural perspective of <country>. Based on <country>'s cultural values, social priorities, and what people from <country> typically find most interesting in their daily lives, rank these TOPICS from MOST to LEAST preferred based on curiosity..''} This task focuses on the contextual preferences of curiosity.
 % to rank topics by their perceived importance or curiosity within that context.
% We then compare these preferences with humans across cultures.

We use several LLMs: GPT-4o, GPT-5, Claude-Sonnet-4, Qwen-3-14b, LLaMA-3-8b-instruct and LLaMA-3-70b-instruct, and average final results across 3 runs.

% - include Task 1 and 2

% - table with a few examples, including topic and country, with human-authored and LLM-generated examples side-by-side

\begin{figure}
\centering
\includegraphics[width=0.8\linewidth]{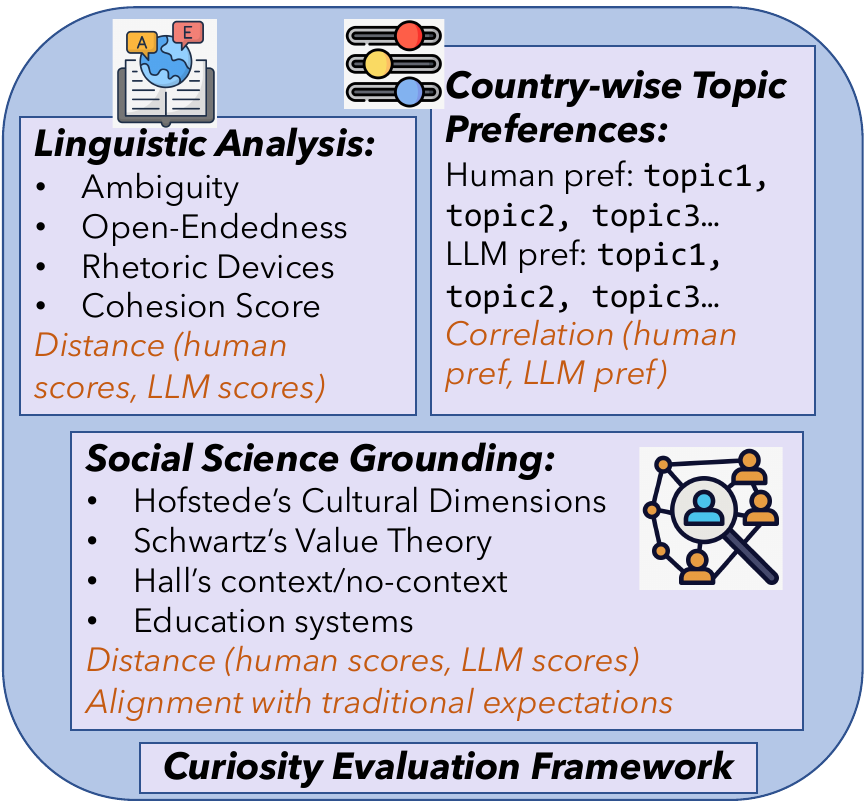}
\caption{\textbf{CUEST Overview:} (1) Linguistic Alignment, (2) Topic Preference Alignment, and (3) Grounding to Social Science Constructs; to compare human- and LLM- questions.}
% \vspace{-20pt}
\label{fig:cuest}
\end{figure}

\section{CUEST: Measuring Human-LLM Alignment}

% brief intro paragraph to state: we measure across three dimensions: linguistic analysis, topics; grounding to social theories
Our framework CUEST proposes three complementary directions to evaluate culture-aware curiosity: \textit{Linguistic Alignment} to capture question styles, \textit{Topic Preference Alignment} to examine content preferences, and \textit{Grounding in Social Science Constructs} to contextualize curiosity within established cultural frameworks. Together, they provide a comprehensive view of how curiosity manifests across cultures. (see Appendix~\ref{sec:qual_questions} for a qualitative analysis of questions). 

% evaluates from a perspective of pre-conceived cultural notions. Here are our top learnings: \newline

\subsection{Linguistic Alignment}
Taking inspiration from previous studies, we look at these four dimensions for linguistic analysis: \textit{(1) ambiguity}: captures the degree to which a question invites multiple interpretations, which is central to curiosity as it leads to exploration~\cite{gruber2019curiosity, tor2020digital, berlyne1960conflict}, \textit{(2) rhetorical devices}: refers to creative or exploratory phrasing that can enhance curiosity expression~\cite{gibbs1994poetics, mcquarrie1996figures}, \textit{(3) open-endedness}: reflects whether a question allows broad answers, similar to curiosity’s drive for knowledge-seeking.~\cite{scialom2019ask, engel2011children, grossnickle2016disentangling}, and \textit{(4) cohesion score}: measures how well a question links concepts together -- connecting ideas can show deeper engagement~\cite{graesser2011computational, crossley2016development}. We provide further details on each metric in Appendix~\ref{sec:metrics}. Note that since we use Google Translate for non-English Yahoo! questions, structural features (open-endedness, cohesion, ambiguity) are largely preserved, though finer nuances like rhetorical devices may not fully carry over; so caution is advised.

\begin{figure}
\centering
\includegraphics[width=\linewidth]{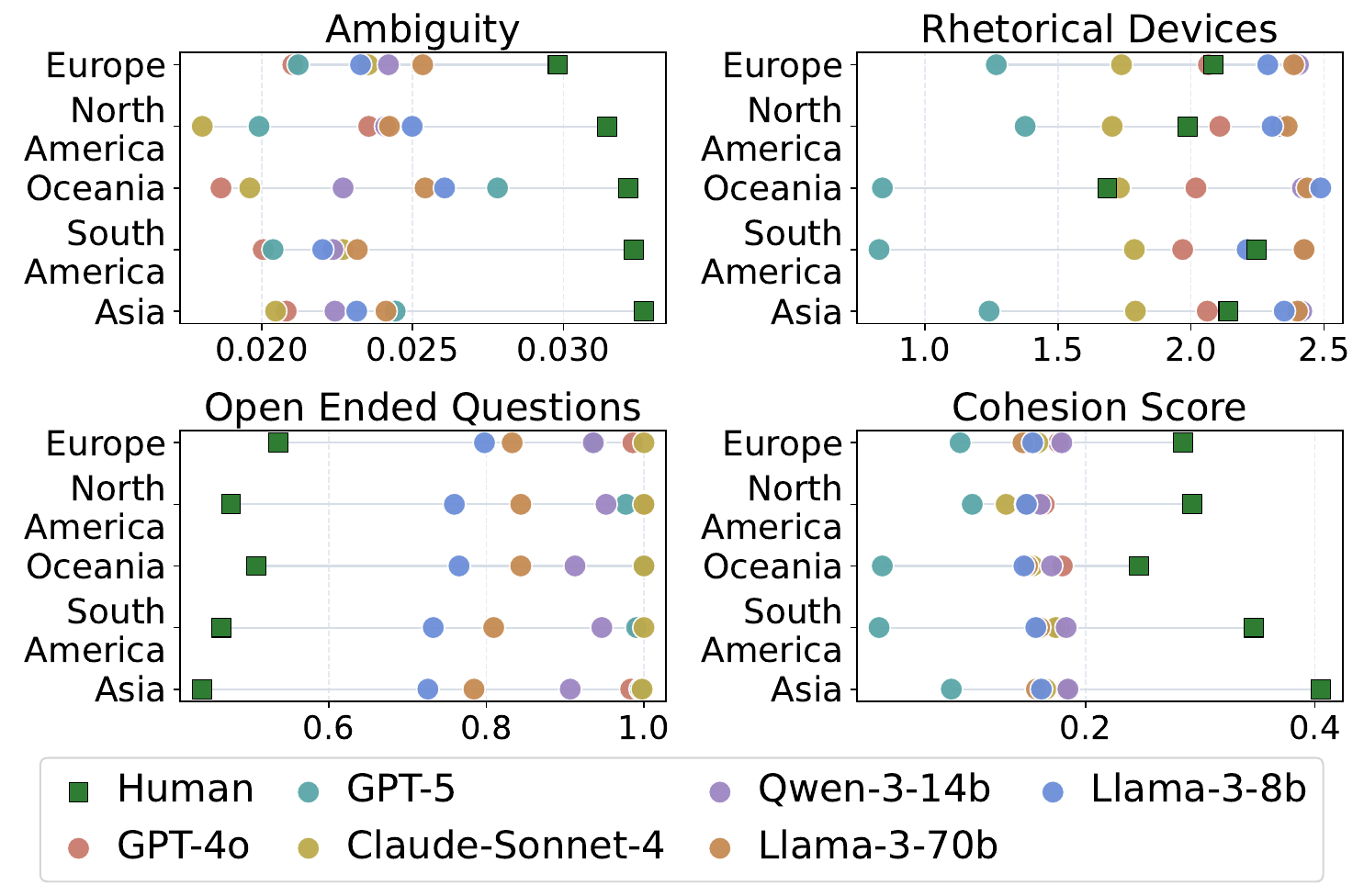}
\caption{\textbf{Linguistic Alignment}. Humans and LLMs diverge: humans high in ambiguity and cohesion score, LLMs high in open-ended questions. }
\vspace{-10pt}
\label{fig:ling_ana_country}
\end{figure}

% \begin{figure}
% \centering
% \includegraphics[width=\linewidth]{latex/images/spider_updated.pdf}
% \caption{Linguistic Analysis for human and LLM-generated questions}
% \vspace{-20pt}
% \label{fig:ling_ana_country}
% \end{figure}

\noindent \textbf{Results.} Figure~\ref{fig:ling_ana_country} shows that humans and LLMs exhibit different curiosity patterns: Humans consistently show a higher cohesion score and ambiguity, whereas LLMs are more open-ended. Rhetorical devices vary across models, with some, like \texttt{LLaMA} variants, are close to human patterns. We should note that rhetorical devices may not fully carry over in translated languages; however, they remain an important metric for assessing curiosity, with notable differences observed among countries. Next, humans standard deviations (0.0785) are higher than LLMs (0.029) (F-stat:\(7.33)\) -- LLMs tend to flatten cross-cultural differences. Finally, we observe only minor continental differences, so we shift to a country-level analysis for finer cultural insights.

\begin{figure}
\centering
\includegraphics[width=\linewidth]{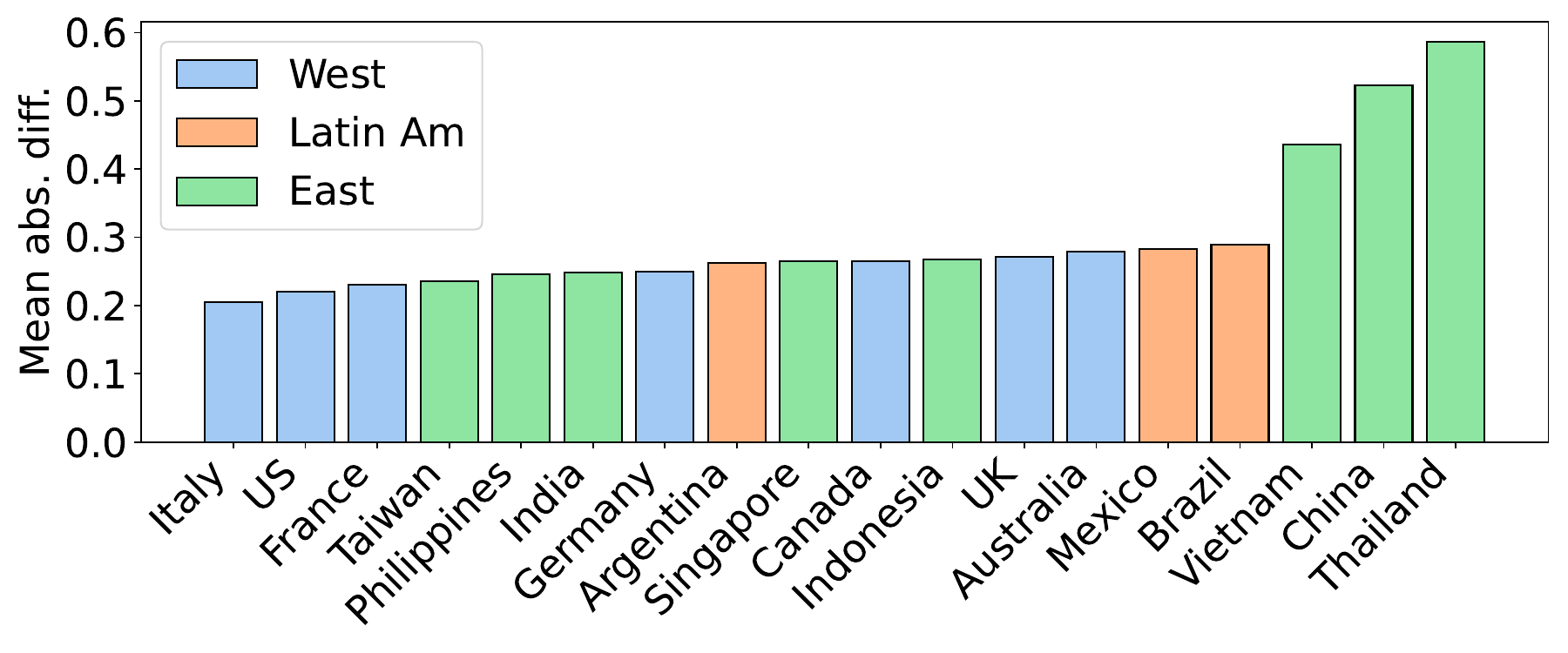}
\caption{\textbf{Country-wise Linguistic Alignment:} (low mean abs diff: high alignment) Western countries show the highest human–model alignment, followed by Latin American and Eastern countries.}
\vspace{-10pt}
\label{fig:regionwise_ling}
\end{figure}

\noindent \textbf{Country-wise analysis.} Fig.~\ref{fig:regionwise_ling} presents the mean absolute differences between models and humans for each country. Lower values mean the model matches closely with human scores. The graph has color-coded regions~\footnote{In our study, all European, North American countries, and Australia are considered as Western, Asian countries as Eastern, and Mexico, Brazil, and Argentina as Latin American} showing a clear pattern - West shows the best alignment; followed by some Eastern and Latin American countries. East Asia, however, shows the largest gaps. 

% (pre-dominantly Europe ($\simeq$0.24) and North America ($\simeq$ 0.26))
% ($\simeq$ 0.28)
% ($\simeq$ 0.35) 
\noindent \textbf{Model wise analysis.} LLaMA-3 models are the closest to human scores overall. GPT-5 has the largest gap, mostly driven by higher differences on open-ended questions and rhetorical devices. Thus, the model rankings (distance in braces) are: \texttt{LLaMA-3-8b} (0.25) > \texttt{LLaMA-3-70b} (0.27) > \texttt{Claude-Sonnet-4} (0.28) = \texttt{GPT-4o} (0.28) > \texttt{Qwen-3-14b} (0.29) > GPT-5 (0.42). 
% It is interesting to see that smaller models perform better, this may arise from (1) more diverse, higher-quality pretraining data in LLaMA models~\cite{dubey2024LLaMA} (however, OpenAI has not fully disclosed the composition and curation criteria), and (2) less aggressive Reinforcement Learning from Human Feedback (RLHF) in smaller models. Whereas larger models undergo more extensive RLHF and safety tuning, which can ``sanitize'' outputs toward being generic, and Western-normative~\cite{aksoy2025whose}. Additional experiments on cultural benchmarks support these findings, as detailed in Appendix~\ref{sec:modelvariation}.

% - move here everything about linguistic alignment - including task def, results, etc.

\subsection{Topic Preference Alignment}

We now compare culture-specific LLM topic preferences with the human-authored ones using Spearman and Kendall correlations ($[-1,1]$) ($1=$ very similar to humans, $0=$ unrelated, $-1=$ inverse).

\begin{figure}
\centering
\includegraphics[width=\linewidth]{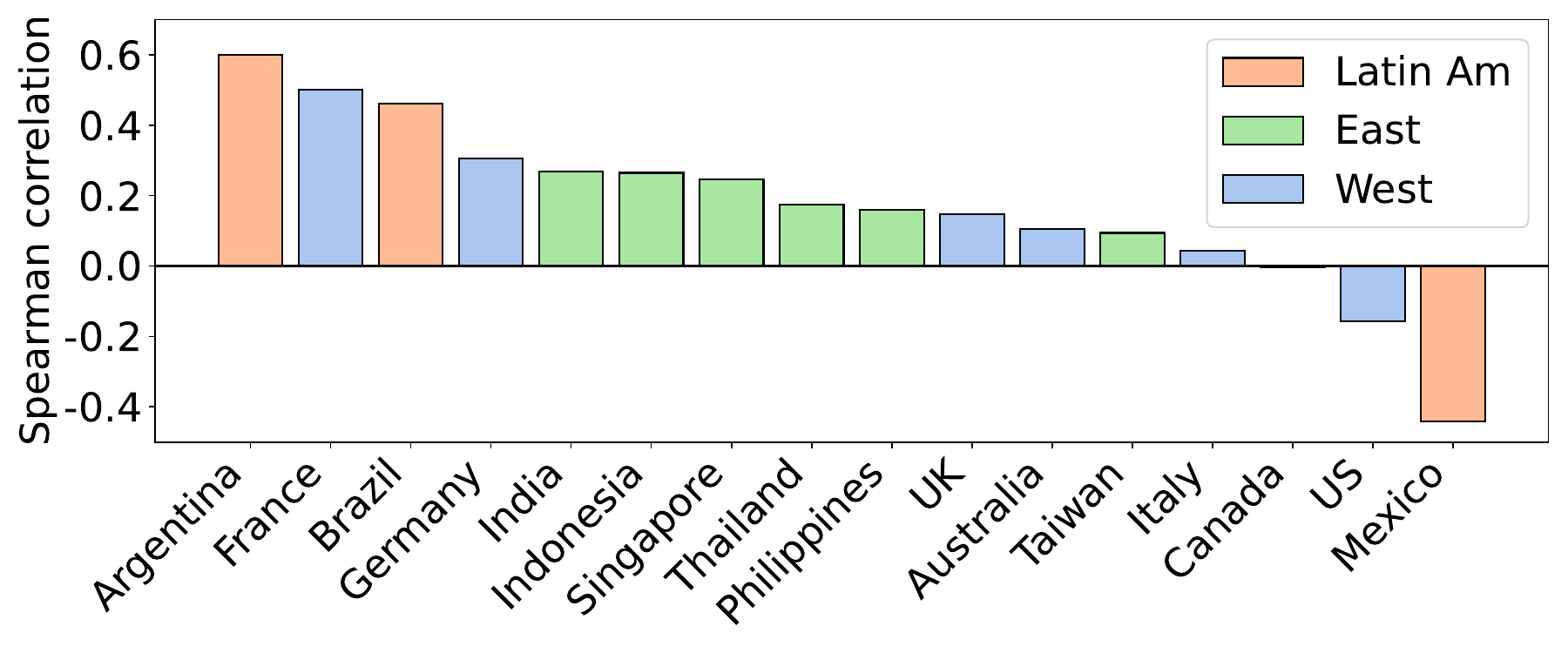}
\caption{\textbf{Country-wise Topic-Preference Alignment}: \texttt{LLaMA-3-8b} correlation with humans. Although we find notable country-level differences, no region-level patterns emerge.}
\vspace{-10pt}
\label{fig:topic_pref_rank_LLaMA}
\end{figure}

% Countries with the highest positive correlations are France, Argentina, Germany, and Taiwan, whereas Mexico and the UK are the countries with the highest negative correlations. We provide the full table . 
% In terms of regions, the closest regions are South America and Europe, whereas the least aligned are North America and Oceania. Asia shows mixed effects (e.g. Taiwan achieving positive correlation in Claude-Sonnet-4, but others show weaker/negative correlations). For the best-performing model, \texttt{LLaMA-3-8B}, we
\noindent \textbf{Country-wise analysis.} 
We compute country-level correlations using the best-performing model: \texttt{LLaMA-3-8b} in Fig.~\ref{fig:topic_pref_rank_LLaMA}. Clear regional differences are not evident: Argentina, France, and Brazil, show the highest correlations, while Mexico and the UK exhibit negative correlations. These findings diverge from our previous results using linguistic analysis, indicating that topic preference captures distinct curiosity patterns. 
% Per-model topic preference correlations are provided in Appendix~\ref{sec:country_topicpref}. 
% We compute country-wise spearman correlation of the best model~\texttt{LLaMA-3-8b} with humans. Fig~\ref{fig:topic_pref_rank_LLaMA} shows the countries correlation with humans. We do not observe clear differences across regions, the top countries are: Argentina, France, Brazil, and Germany, while Mexico and the UK show negative correlations. 
% These findings diverge from our previous results using linguistic analysis, indicating that topic preference and linguistic style capture distinct curiosity styles. 
% It is also particularly interesting given that most LLMs are predominantly trained on WEIRD data distributions, where we would expect North America to have higher alignment. 

% \noindent \textbf{Understanding East-West-Lat Am. divergences.} 
We qualitatively analyze topic preferences to examine country-wise differences. We observe that alignment is strongest when human curiosity aligns with cultural stereotypes, for example, Argentina and Brazil prefer family and entertainment, consistent with collectivist values~\cite{hofstede2001culture,krys2022outside} and entertainment-rich online cultures\footnote{\href{https://latamjournalismreview.org/articles/content-creators-and-online-video-gain-ground-over-traditional-media-in-latin-america-new-report-finds/\#:\~:text=\%E2\%80\%9CDependence\%20on\%20social\%20media\%20is,\%E2\%80\%9D}{\textit{Online video surpasses traditional media in Latin America, new report finds}}}. Similarly, France and Germany highlight arts and humanities, reflecting cultural history (e.g., European Wikipedias~\cite{ahmed2023representation}). Divergence grows when curiosity deviates from stereotypes or focuses on underrepresented topics: in the UK, where humans prioritize arts over politics despite the dominant political discourse (e.g., Brexit). Overall, while LLMs align closely with Western contexts, they also reinforce Western stereotypes~\cite{borah-etal-2025-towards}. Detailed country-level preferences per model are in Appendix~\ref{sec:country_topicpref}.
% We also perform a qualitative analysis of topic preferences to better understand country-wise differences.We observe strongest alignment when human curiosity overlaps with topics that align with traditional stereotypes. For instance, in Argentina and Brazil, both prioritize family and entertainment—consistent with collectivist values~\cite{hofstede2001culture,krys2022outside} and abundant online entertainment content~\cite{obo}\footnote{https://tinyurl.com/3ky9zaw3}. Similarly, in France and Germany, emphasis on arts and humanities mirrors cultural history and European Wikipedias\footnote{https://tinyurl.com/ys9a5yk3}. Conversely, divergence is higher when human curiosity defies stereotypes or focuses on underrepresented lifestyle topics. For example, in the UK, humans favor \textit{arts and humanities} over politics despite the abundance of political discourse (e.g., Brexit). Thus, while LLMs closely align with Western contexts, they also strongly embed Western stereotypes~\cite{borah-etal-2025-towards}. Detailed country-level topic preferences are in Appendix~\ref{sec:country_topicpref}. 
% - move here everything about topic alignment - including task def, results, etc.

\begin{table}[h!]
\centering
\small
\begin{tabular}{p{2cm}p{2cm}p{2cm}}
\toprule
\textbf{Model} & \textbf{Spearman} & \textbf{Kendall} \\
\midrule
\texttt{LLaMA-3-8b} & 0.17 & 0.13 \\
\texttt{LLaMA-3-70b} & -0.09 & -0.04 \\
\texttt{Qwen-3-14b} & -0.16 & -0.13 \\
\texttt{Claude-Sonnet-4} & -0.12 & -0.08 \\
\texttt{GPT-4o} & -0.29 & -0.20 \\
\texttt{GPT-5} & -0.22 & -0.14 \\
\bottomrule
\end{tabular}
\caption{\textbf{Topic Pref. Alignment:} Model-Human Corr. Scores: We find LLaMA variants have the highest correlation while GPT variants have the lowest. }
\vspace{-10pt}
\label{tab:model_corr}
\end{table}

\begin{figure}
\centering
\includegraphics[width=\linewidth]{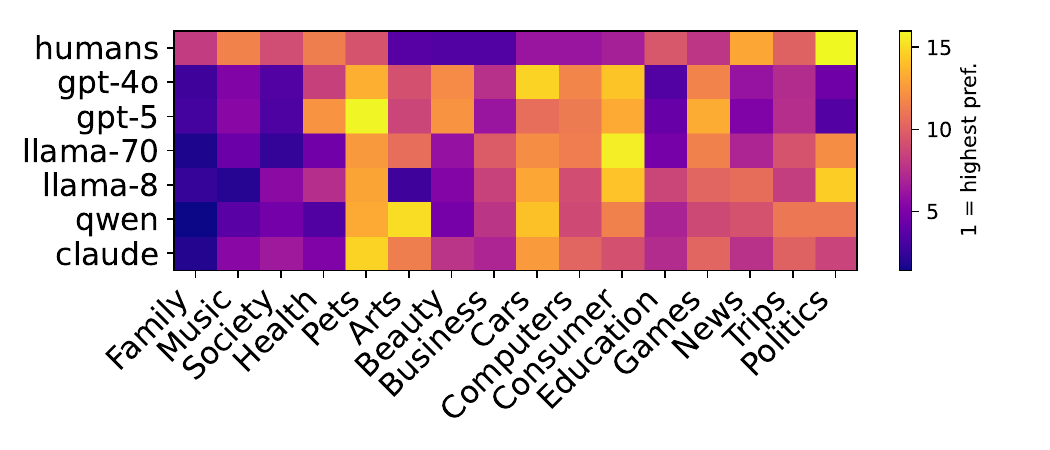}
\caption{\textbf{Topic-Preference Ranking}: Heatmap of preferred topics (shortened) averaged across countries for human and models: (1- highest preferred). It shows a clear divergence between humans and LLMs.}
\vspace{-10pt}
\label{fig:topic_pref_ranks}
\end{figure}

\noindent \textbf{Model-wise analysis.} Table~\ref{tab:model_corr} shows that only \texttt{LLaMA-3-8B} achieves a modest positive correlation; while all other models diverge, with \texttt{GPT-4o} exhibiting the strongest negative correlation. 
Figure~\ref{fig:topic_pref_ranks} visualizes average topic preferences. LLMs consistently emphasize `Family and relationships', `Music and entertainment', etc. whereas humans prioritize `Arts and humanities', `Beauty and style', and `Business and finance', highlighting systematic differences in curiosity-driven focus.

\subsection{Grounding to Social Science Theory}
\label{social}
As discussed earlier, grounding in social science constructs is essential to ensure comparison to prior cross-cultural studies and established cultural patterns. Specifically, we focus on the following:

\noindent \textbf{1) Hofstede's Cultural Dimensions}: We operationalize 4 dimensions related to curiosity: \textit{(1) Uncertainty Avoidance (UA):} comfortability with ambiguity and risk: high UA often discourages exploratory questioning and low UA allows higher curiosity~\cite{hofstede2001culture, kashdan2009curiosity}, \textit{(2) Individualism-Collectivism (IC):} expressing as a self-related inquiry or a social exploration~\cite{triandis1995multimethod}, \textit{(3) Indulgence-Restraint (IR):} indulgent culture frame curiosity as playful and hedonic, while restrained cultures emphasize discipline and self-control~\cite{hofstede2010cultures}, and \textit{(4) Long-/Short-term orientation (LSO):} if curiosity is future oriented or present/past-oriented and pragmatic~\cite{hofstede2010long}. \newline
        \noindent \textbf{2) Schwartz Cultural Values}: We focus on \textit{Openness to Change versus Conservation}. Openness values novelty, questioning, and self-direction, directly promoting curiosity as a motivational force~\cite{schwartz1992universals, kashdan2009curiosity}. But, conservation emphasizes conformity, and tradition -- constraining curiosity to ``safe'' domains. \newline
        % Schwartz identifies several dimensions, such as Embeddedness (tradition, hierarchy) vs Autonomy (individual freedom), Hierarchy (inequality) vs Egalitarianism (equal rights), Mastery (control over environment) vs Harmony (fitting in), as important in reflecting cultural values of different countries. Cultures emphasizing embeddedness and hierarchy favor formal, layered phrasing (higher sentence complexity) and higher Flesch–Kincaid scores, reflecting power and conservatism, while cultures valuing autonomy and openness to change produce lower-grade, more accessible text. Open-endedness aligns with self-direction and openness, promoting exploration, creativity, and questioning.
        \noindent \textbf{3) Halls' Contextual dimensions}: Hall's \textit{high context cultures} rely on implicit cues, and tentativeness, meaning curiosity often manifests in indirect forms. In contrast, \textit{low-context cultures} are explicit, detailed, and analytical, with curiosity expressed through direct questions~\cite{hall1976beyond}. \newline %gudykunst1988culture
        % Hall distinguishes High-Context culture, requiring implicit communication and collectivist culture, from Low-Context culture, requiring explicit communication and individualistic culture. High-context cultures favor indirect communication, leading to higher sentence complexity and more formal, elaborate structures, which increase reading difficulty. Low-context societies encourage direct and explicit questioning, resulting in more exploratory and open-ended queries, while high-context cultures rely on implicit or less overtly open-ended forms.
        \noindent \textbf{4) Education systems} -- \textit{Rote vs. Holistic Learning}. Rote-learning styles are rule-based, authority-driven, and focused on correct answers~\cite{biggs1996western}, and holistic learning styles are exploratory and open-ended~\cite{nisbett2010geography}. 
        % Education systems differ significantly across countries, with some emphasizing exam-driven approaches and others prioritizing holistic learning. For instance, Eastern countries like India and Taiwan often center around academic rigor and performance, while Western countries like France and the United States focus on broader, well-rounded curricula.
    % \end{enumerate}

\begin{figure}
\centering
\includegraphics[width=0.8\linewidth]{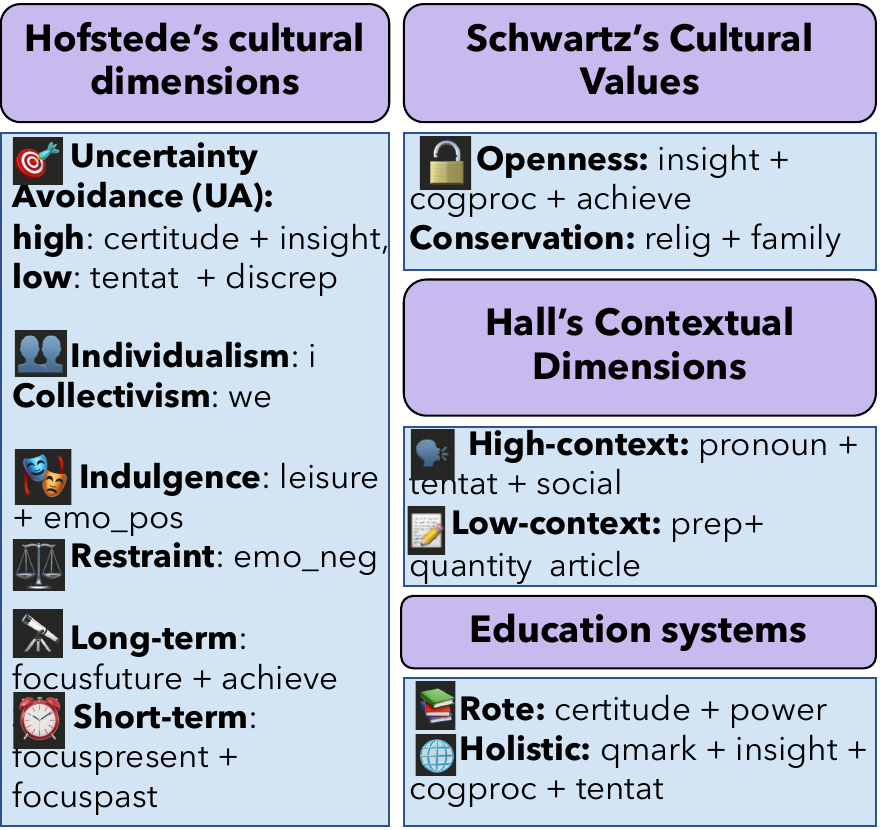}
\caption{\textbf{Social Science Constructs:} Computation using LIWC dimensions.}
\vspace{-10pt}
\label{fig:liwc_ss}
\end{figure}

% \noindent \textbf{Results.} In this section, we focus on the differences between human and LLMs in the context of the above theories and constructs. 
\noindent We compute LIWC scores corresponding to the above dimensions, showed in Fig~\ref{fig:liwc_ss}.

%to humans. 
%, whereas \texttt{GPT-5} is the most divergent. 

\begin{figure}
\centering
\includegraphics[width=\linewidth]{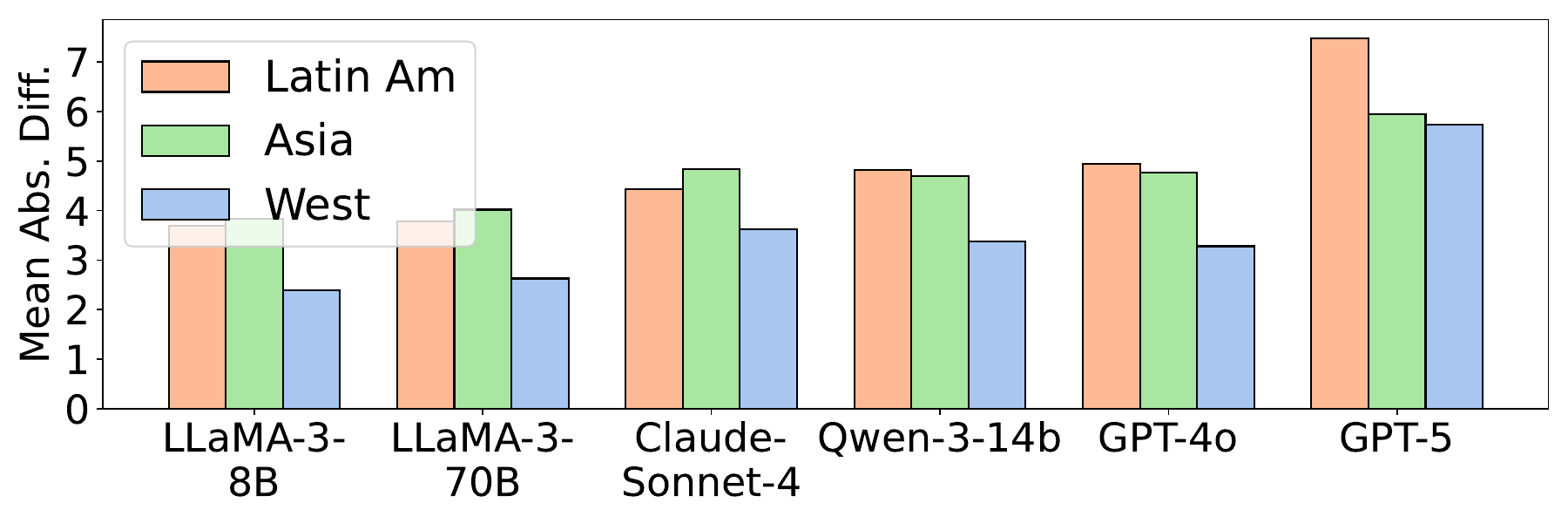}
\caption{\textbf{Social Science Theory Alignment:} Region-wise MAE for models shows West vs East/Lat Am. Differences are prominent across models.}
\vspace{-10pt}
\label{fig:region_ss}
\end{figure}

\noindent \textbf{Region-wise analysis.} Fig~\ref{fig:region_ss} shows the mean absolute differences between model and human using LIWC scores. All models are consistently closer to humans in Western/Anglo countries than in Eastern or Latin American countries. 
% \texttt{LLaMA} models are closest to humans reducing regional gaps, while \texttt{GPT-5} shows the largest. 

\noindent \textbf{Model-wise analysis.} Taking human-authored questions as baseline, we find the following order of models from closest to farthest: LLaMA-3-8b > LLaMA-70B > Claude > Qwen > GPT-4o > GPT-5. Thus, \texttt{LLaMA-3-8b} is again the best-aligned model. We also perform theory-wise analysis (detailed in Table~\ref{tab:model_theories} and Appendix~\ref{sec:social_comp_model}). 

\definecolor{darkgreen}{RGB}{0,100,0}
\begin{table*}[h!]
\centering
\small
\begin{tabular}{p{8cm}p{1cm}p{1.9cm}p{1cm}p{0.9cm}p{0.8cm}}
\toprule
\textbf{Theory and Expectation} & \textbf{Human}  & \textbf{LLaMA-3-8b}  & \textbf{Claude}  & \textbf{GPT-5} & \textbf{Qwen} \\
\toprule
\textbf{IC (Hofstede)} – Western societies should score higher in Individualism and Asian/Latin societies score higher in Collectivism & \textcolor{orange}{Mixed} &  \textcolor{darkgreen}{Aligned} & \textcolor{red}{Not aligned} & \textcolor{orange}{Mixed}  & \textcolor{orange}{Mixed}  \\ \midrule
\textbf{UA (Hofstede)} – Mexico, Brazil and other emerging economies should have higher scores, while Anglo countries are lower & \textcolor{orange}{Mixed} & \textcolor{red}{Not aligned} & \textcolor{red}{Not aligned} & \textcolor{orange}{Mixed} & \textcolor{orange}{Mixed} \\ \midrule
\textbf{IR (Hofstede)} – U.S., Australia and other Western societies should be most indulgent &  \textcolor{darkgreen}{Aligned} &  \textcolor{darkgreen}{Aligned} & \textcolor{red}{Not aligned} & \textcolor{orange}{Mixed} & \textcolor{orange}{Mixed} \\ \midrule
\textbf{LSO (Hofstede)} – Asian cultures emphasize the future more (long-term) than Western cultures (short-term) & \textcolor{orange}{Mixed} & \textcolor{orange}{Mixed} & \textcolor{red}{Not aligned} & \textcolor{orange}{Mixed} & \textcolor{orange}{Mixed} \\ \midrule
\textbf{Openness vs conservation (Schwartz)} – Collectivist societies emphasize tradition and family (conservation), Western societies emphasize openness &  \textcolor{darkgreen}{Aligned} &  \textcolor{darkgreen}{Aligned} & \textcolor{orange}{Mixed} & \textcolor{orange}{Mixed} & \textcolor{orange}{Mixed} \\ \midrule
\textbf{High vs low context (Hall)} – Asia and Latin America are high‑context, Western cultures low‑context & \textcolor{red}{Not aligned} &\textcolor{red}{Not aligned} & \textcolor{red}{Not aligned} & \textcolor{red}{Not aligned} & \textcolor{red}{Not aligned} \\ \midrule
\textbf{Rote vs holistic learning} – Most East Asian countries have been characterized as rote-oriented, whereas Western education systems are often characterized as more holistic, consisting of critical thinking, creativity, emotional and social development & \textcolor{orange}{Mixed}  &\textcolor{orange}{Mixed}  &\textcolor{orange}{Mixed}  &\textcolor{orange}{Mixed}  &\textcolor{orange}{Mixed}  \\ 
\bottomrule
\end{tabular}
\caption{\textbf{Alignment of humans and LLMs to traditional expectations} across social science theories and constructs}
\label{tab:traditional}
\vspace{-10pt}
\end{table*}

\noindent \textbf{Alignment of Humans and LLMs to traditional expectations.}
The traditional literature on intercultural psychology expects differences across countries based on the above constructs. Table~\ref{tab:traditional} shows whether the humans and models follow traditional expectations (\textcolor{darkgreen}{Aligned}), depart from them (\textcolor{red}{Not aligned}), or show no clear pattern except a few matches (\textcolor{orange}{Mixed/Partially aligned}). Humans themselves do not consistently follow the traditional expectations. This may be because: \textit{(1)} Yahoo is an anonymous platform, effectively reducing normative pressure~\cite{guo2021anonymity, ellison2016question} \textit{(2)} cultural evolution and globalization leading to shifts in behaviors over time~\cite{witte2020new, kittler2011special, greenfield2016social}. 

Among models, \texttt{LLaMA-3-8b} has the highest alignment. \texttt{Claude-Sonnet-4} exhibits the most inverted patterns, while \texttt{GPT-5} and \texttt{Qwen} display mixed effects. 
Theory-wise, both human and models align on IR and Schwartz’s dimensions, with \texttt{LLaMA-3-8B} also aligning on IC. Hall’s high–low context shows the strongest non-alignment, while education systems yield the most mixed outputs. 

In all experiments, \texttt{LLaMA-3-8b} performs the best, showing the superiority over larger models. This may be due to: (1) diverse, higher-quality pretraining data~\cite{dubey2024llama} (though OpenAI has not fully disclosed their data), and (2) less aggressive RLHF, whereas larger models undergo heavier RLHF and safety tuning, which can ``sanitize'' outputs toward generic, Western-normative patterns~\cite{aksoy2025whose}. Additional experiments in Appendix~\ref{sec:modelvariation} support this.

\section{Inducing Curiosity in LLMs}

Having found interesting culture-specific curiosity differences in humans and LLMs, we now shift our focus to inducing culture-aware curiosity in models. We focus on countries where human-LLM alignment was low -- Brazil, the UK, and the Philippines (from Lat Am., Western, and Eastern regions, respectively), and each with sufficient data.
% we now shift our attention to reducing the human-LLM gap in curiosity by taking a few countries that previously showed low alignment (using the evaluation framework) with humans per region. We take Brazil, the UK, and the Philippines from the Latin America, West, and East regions. 

\noindent \textbf{Datasets.} We use two datasets: \textbf{(1) Yahoo! Answers}, and \textbf{(2) Reddit}. Since data size per country is limited, we apply simple lexical augmentation techniques (e.g., synonym replacement, word swaps) while preserving the overall style of the questions, resulting in 1,000 examples per country. For fine-tuning, we solely use the augmented data from \textit{Yahoo!}, whereas for \textit{Reddit}, we collect questions from three country-specific subreddits: \texttt{r/brazil}, \texttt{r/askuk}, and \texttt{r/philippines}. 
% \textit{NormAd} is a benchmark of culturally grounded social norms from 75 countries using the \texttt{LLaMA-3} variants. 

\noindent \textbf{Methodology.} We fine-tune \texttt{LLaMA-3-8b} that showed the closest alignment with humans, using two approaches: full fine-tuning and adapter-based fine-tuning. We use two objectives: (1) \textit{obj1}: train models to generate questions given instruction: ``Write one question that could come from a person belonging to <x>''; and (2) \textit{obj2}: augment each dataset question with a preceding conversational statement generated by \texttt{GPT-4o} (e.g., for a question: ``Which book was the spag yeti in?'', GPT-4o generates a preceding conversation: ``I was reminiscing about those quirky children's books we used to read to our daughter…''). This format encourages the model to learn natural, curiosity-driven questioning. We test both country-specific and combined all-country settings, and three data settings: \textit{yahoo-only}, \textit{reddit-only} and both \textit{yahoo+reddit}. 
% Our approach (2) - using lightweight LoRA adapters enables us to capture culturally grounded curiosity without the expense of full fine-tuning, thereby reducing computational costs.

\definecolor{pastelAdapter}{HTML}{A3C4F3}   % soft blue
\definecolor{pastelFullFT}{HTML}{FFB3C1}    % soft rose
\definecolor{pastelPrompt}{HTML}{CDEAC0}    % soft green
\definecolor{obj1}{HTML}{F39C12}           % orange
\definecolor{obj2}{HTML}{16A085}           % teal
\definecolor{yahoo}{HTML}{7B68EE}
\definecolor{reddit}{HTML}{2E8B57}
\definecolor{yahooreddit}{HTML}{C04040}
\definecolor{countryAll}{HTML}{1F77B4}
\definecolor{countrySpecific}{HTML}{8C564B}
\definecolor{txtColor}{HTML}{222222}

\begin{figure}
\centering
\includegraphics[width=\linewidth]{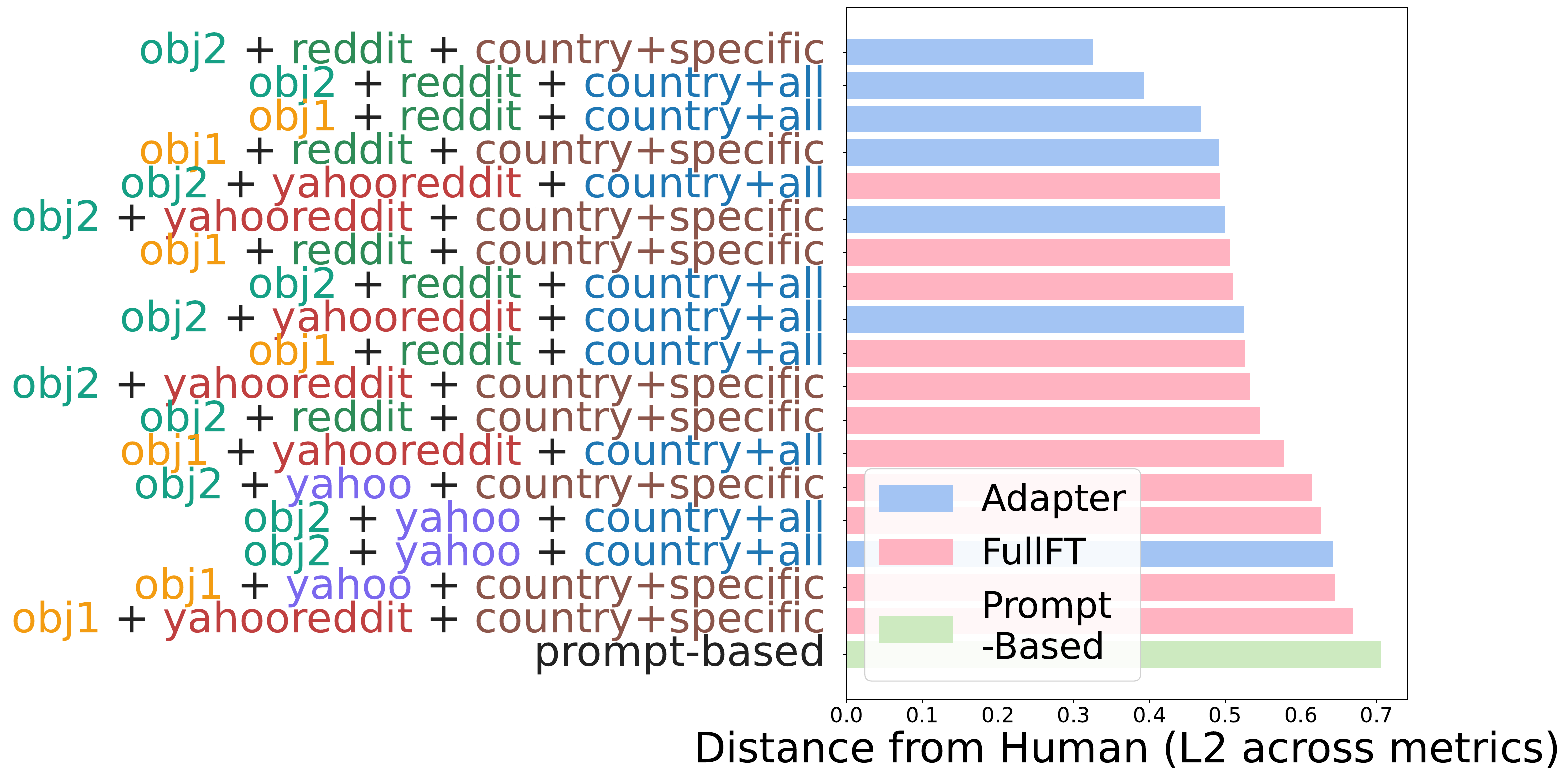}
\caption{\textbf{Linguistic Analysis on fine-tuned vs non-fine-tuned models}: y-axis shows configs, while the x-axis show model-human distances (L2) (averaged across metrics and countries). \colorbox{pastelAdapter}{adapter}-based methods with \textcolor{obj2}{obj2} on \textcolor{reddit}{Reddit} perform the best, while prompt-based methods performing poorly. }
\vspace{-20pt}
\label{fig:avg_diff_ft}
\end{figure}

\noindent \textbf{Evaluation.} We use two approaches: \textbf{(1) culture-aware curiosity}: using CUEST -- comparing model outputs with humans on the Yahoo dataset  and \textbf{(2) inherent curiosity}: to determine models’ \textit{inquiry ability} in conversational settings. For (2), we design 40 generic statements and 30 cultural statements (specific to the three countries), that are open-ended/ambiguous and naturally invite follow-up questions (some examples with LLM outputs provided in Table~\ref{tab:curious_induce_inherent}, full dataset in Appendix~\ref{sec:dataset_build}). To evaluate, we use: (1) \textit{Curiosity Rate} -- does the model ask information-seeking question?  and (2) \textit{Relevance} -- is the question relevant to the statement? Evaluation is performed by two human evaluators over one conversation round. (see Appendix~\ref{sec:ann} for annotation details). 

% (3) \textit{Over-/under-curiosity} -- Do models consistently ask questions even in situations that do not require a ?
% \footnote{Note that we do not assess cultural appropriateness in these metrics, as our focus is on inherent model curiosity rather than culture-specific reasoning.}
% For both datasets, we apply simple lexical augmentation techniques (e.g., synonym replacement, word swaps) while preserving the overall style of the questions, resulting in 1,000 examples per country. For fine-tuning, we solely use the augmented data from Yahoo!, whereas in the case of Reddit, we collect questions from three country-specific subreddits: \texttt{r/brazil}, \texttt{r/askuk}, and \texttt{r/philippines}. Fine-tuning is conducted separately for each dataset using country-specific data, and 

\begin{table*}[htbp]
\centering
\small
\begin{tabular}{p{2.8cm}|p{2.4cm}|p{4.6cm}|p{4.4cm}}
\toprule
\textsc{Statement} & \textsc{Adapter} & \textsc{Full-FT} & \textsc{Non-FT-Prompt-Based} \\
\midrule
\multicolumn{4}{c}{\textit{Everyday statements}} \\
\midrule
A guest walked straight into the kitchen at someone's house. &
Is this a sign of trust? Or is it a sign of something else? 
& The scenario of a guest walking straight into the kitchen at someone's house may have different implications ....
& In many Western cultures, it's considered impolite to enter a kitchen without being invited, however, Asian ... \\
% \midrule
% Someone talked to their plants every morning. &
% How do you talk to plants? 
% & As a curious agent, I would say that this scenario has implications for our understanding of human-animal relationships and the boundaries between humans and non-human ....
% & As a curious agent, I'd say that this scenario has cultural and possibly therapeutic implications. Talking to plants may be a form of mindfulness practice, ....  \\
\midrule 
\multicolumn{4}{c}{\textit{Culture-based statements (country)}} \\
\midrule 
Football matches in Brazil usually come with unique social rituals & What are some popular traditions in Brazilian football? & As a curious agent, I would like to better understand the social rituals and how they contribute to the country's vibrant culture... &  As a curious agent, I'm intrigued by the vibrant cultural practices surrounding football in Brazil, where fans often engage in chanting, ... \\
%  \midrule
% In UK workplaces, meetings often begin with a particular social convention.	& What is the most common icebreaker? & 
% In the UK, it is a common practice to start meetings with a brief "small talk" or "chit-chat" session, where attendees ... & 
% As a curious agent, I'm intrigued to learn that in UK workplaces, meetings often start with a brief moment of silence or a nod to the Queen's Speech.. \\
\bottomrule
\end{tabular}
\caption{\textbf{Inherent curiosity evaluation:} Examples from three model responses to benchmark statements.}
\label{tab:curious_induce_inherent}
\vspace{-10pt}
\end{table*}
% s (two-trained: adapter- and full-finetuned(full-ft), non-ft-prompt-based)
% The y-axis shows model settings (20 out of 24 total configs), and the x-axis shows L2 distances between models and humans. 
\noindent \textbf{Results - Culture-aware Curiosity.} Fig.~\ref{fig:avg_diff_ft} presents \textit{linguistic analysis} results and shows that fine-tuning consistently narrows stylistic gaps. Reddit-trained models are closest to humans, likely due to exposure to diverse rhetorical strategies, while Yahoo-trained models show only modest gains, maybe due to the use of lexically augmented data for fine-tuning. Adapter-based models outperform full fine-tuning, likely because adapters add curiosity-driven behaviors without overwriting the general flexibility of the base model, whereas full fine-tuning may over-specialize to the training format. Finally, \texttt{obj2} does better than \texttt{obj1}, maybe due to added conversational scaffolding.

Topic preference analysis shows that only \texttt{obj2+reddit+country\_all} (0.43/0.31) and \texttt{obj2+yahooreddit+country\_all} (0.32/0.27) exceed the prompt baseline (0.26/0.19), leading to closer (Spearman/Kendall) correlation with human preferences. Results are averaged across three countries, with no notable differences between country-specific and country-all training. These trends are similar linguistic analysis, with Reddit-adapter-obj2-trained models performing the best. Linguistic analysis benefits substantially more from fine-tuning than topic preference analysis. Further details are provided in Appendix~\ref{sec:ft}.

% We also perform topic preference analysis using the fine-tuned models -- only two models have higher correlation than the initial prompt-based analysis - obj2+reddit+country\_all (0.43/0.31) and obj2+yahooreddit+country\_all (0.32/0.27), showing that reddit trained, and obj2 trained models show similar content preferences to humans. The prompt based analysis again has a modest correlation of (0.26/0.19). Note that this is averaged across the three countries. In addition, we do not find any significant differences when we train specifically to country or country-all. Therefore, observations are similar for both linguistic and topic preference analysis. However, several fine-tuned models achieve better results for linguistic analysis than topic preference. 

% Fine-tuning (full and adapter-based) consistently narrows the gap, with the smallest gap in Semantic Ambiguity, suggesting models capture ambiguity in ways closer to humans. Reddit-trained adapters align most closely with human performance on Rhetorical Devices and Open-Ended Questions, likely due to exposure to diverse rhetorical strategies and informal styles. Yahoo-trained adapters also improve but less, maybe due to use of only lexically augmented data for fine-tuning. Full fine-tuning performs the best for Cohesion but does not do very well on other metrics. Country-level results are in Appendix~\ref{sec:ft}. \angana{add topic based analysis}

\noindent \textbf{Results - Inherent Curiosity.}
% - current sec 3
% For inherent curiosity, we only utilize the country-all models, as this is a general purpose task. Additionally, we choose reddit-trained and yahoo-reddit trained models with obj2 (as this objective aligns better with the task requirements). We also do experiments with obj1, details are in Appendix~\ref{sec:}.  
We use country-all models for inherence curiosity, as it tests general-purpose LLM inquiry capabilities. We choose the best-performing configurations from each training type: \textit{adapter+obj2+reddit+country+all}, fullft+obj2+yahooreddit+country+all and \textit{prompt-based}. Note that the prompt used for inference differs from training: \texttt{``You are a CURIOUS agent focused on understanding and creating cultural awareness.''} without explicit instructions to ask questions. This allows for evaluating inherent question-asking ability. We find that adapter-ft models pose most questions (75\%), whereas the full-ft models hardly ask any questions (\(20\%\)). 
% and provide the statement, 

Qualitative analysis shows that full-ft model inquiries often start with \texttt{``I am interested in…''} and continue with statements rather than explicit questions, reflecting a form of \textit{diversive implicit curiosity}~\cite{berlyne1960conflict}. This maybe due to training differences: Adapters preserve base model priors and enhance questioning, yielding better generalization~\cite{pfeiffer2023modular}, while full fine-tuning may reduce flexibility under inference prompt shifts (as found in previously). Finally, prompt-only models fail to generate any questions. Relevance remains high for all cases (98–100\%). All the above results are corroborated by a strong annotator agreement (\(90.21\%\): curiosity rate and \(81\%\): relevance). We also perform a no-country evaluation in cultural question in Appendix~\ref{sec:nocountry_inherent}.
% , which we leave for future research.
% Qualitative analysis reveals that most model inquiries begin with \texttt{``I would be interested to learn more…''} and proceed with extended statements rather than ending with explicit questions. This results in vague information-seeking behavior, reflecting a form of diversive curiosity~\cite{berlyne1960conflict}, which we leave for future research. This maybe due to training differences: adapters preserve instruction-following priors of the base model, adding lightweight adjustments that enhance question-asking behavior. However, full-finetuning rewrites the core parameters, which may reduce flexibility when inference prompts deviate from the training. Prior work has shown that adapters often yield better generalization and robustness under distribution shifts compared to full-finetuning~\cite{pfeiffer2023modular}. Finally, the prompt-only (non-finetuned) model fails to generate any questions despite being instructed to act curious. In terms of relevance, all models score high (~98–100\%). All the above results are corroborated by a strong annotator agreement (\(90.21\%\) for curiosity rate and \(81\%\) for relevance).

% We focus on \texttt{reddit-} and both \texttt{yahoo-reddit–trained} models with \texttt{obj2} (which aligns more closely with the task requirements). Experiments with \texttt{obj1} are included in Appendix~\ref{sec:}. Thus, we provide results for two settings - \texttt{obj2 + }
% All the above datasets capture complementary aspects of cultural adaptability. 

\subsection{Downstream Impact of Curious LLMs}
% These downstream task requires models to judge whether a short scenario is socially acceptable within a given cultural context.
To further assess the functional value of culture-aware curiosity in LLMs, we extend our evaluation to three benchmarks that capture complementary aspects of cultural understanding: (1) NormAd~\cite{rao-etal-2025-normad} contains culturally-grounded social situations across 75 countries and tests models’ judgments of social acceptability (we restrict to binary (acceptable vs.\ not acceptable) and exclude neutral); (2) CulturalBench~\cite{chiu2024culturalbench} is a set of 1,227 human-verified questions from 45 regions across different topics to probe cultural understanding (we use the Hard subset); and (3) Cultural Commonsense~\cite{shen-etal-2024-understanding} tests models’ ability to reason about culturally specific everyday inferences (we use the English fill-in-the-blanks subset on country customs)(See Table~\ref{fig:bench} for examples). These tasks demonstrate why curiosity is critical: instead of defaulting to Western assumptions, a model with culture-aware curiosity would ask questions to better interpret contexts.
% We input country information in the prompt for inference. 

In our experiments, we compare a direct baseline (models classify immediately) with two curiosity-enabled variants: (a) prompting models to generate one or two sub-questions per prompt and also answer them before deciding (inspired by Self-Ask~\cite{press2022measuring}), and (b) an integrated approach combinding the best-performing fine-tuned model from previous experiments to generate questions and base model for answers and decision.

\begin{table}[htbp]
\centering
\small
\begin{tabular}{p{3cm}p{1cm}p{1cm}p{1cm}}
\toprule
\textsc{Condition} & \textsc{NA} & \textsc{CB} & \textsc{CS} \\
\midrule
non-curious & 70.48\% & 64.71\% & 48.48\%  \\ 
curious (prompt-ask) & \colorbox{pastelteal}{72.09\%} & 67.64\% & 49.64\% \\
curious (ft+prompt-ask) & 71.06\% & \colorbox{pastelteal}{68.21\%} & \colorbox{pastelteal}{56.16\%} \\
\bottomrule
\end{tabular}
\caption{\textbf{Downstream tasks (\% acc):} NormAD (NA), CulturalBench (CB), and Cultural Common-Sense (CS)}
\label{tab:normad}
\vspace{-10pt}
\end{table}

% with the country-informed settings yielding the best performance. 
\noindent \textbf{Results.} Table~\ref{tab:normad} shows that curiosity-enabled variants outperform the non-curious baseline. Notably, the \texttt{ft+prompt-ask} setting -- trained on unrelated Reddit data outperforms \texttt{prompt-ask} overall, showing that models trained to inherently ask questions can improve cultural adaptability regardless of the data. Overall, curiosity helps models avoid overgeneralization and better grasp context (see Appendix~\ref{sec:downstream_further} for further fine-tune experiments). Finally, we provide a qualitative analysis of model-generated follow-up questions in Appendix~\ref{sec:qual_followup}. These findings strengthen our claim that culture-aware curiosity is a practical mechanism for improving cultural adaptability in LLMs.

% \texttt{prompt+ft} mode performs better than simple prompting - showing that a model trained to ask questions on a completely different dataset (reddit and countries). This indicates that curiosity assists models in avoiding overgeneralization and understanding the context better.  These findings strengthen our claim that LLMs should be equipped with culture-aware curiosity to improve their adaptability across diverse cultural contexts. Finally, we perform a qualitative analysis on the follow-up questions asked by models, detailed in Appendix~\ref{sec:qual_followup}. 
% In turn, these findings strengthen our claim that culture-aware curiosity is a practical mechanism for aligning models with nuanced, culturally variable human behavior.

% \begin{table}[htbp]
% \centering
% \small
% \begin{tabular}{lcc}
% \toprule
% \textbf{Condition} & \textbf{LLaMA-3-8b} & \textbf{LLaMA-3-70b} \\
% \midrule
% non-curious & 63.20\% & 65.73\% \\
% non-curious + country & 70.48\% & 69.58\% \\ 
% curious & 67.49\% & 79.21\% \\
% curious + country & 72.09\% & 81.04\% \\
% \bottomrule
% \end{tabular}
% \caption{NormAD Performance (Acc \%)}
% \label{tab:normad}
% \end{table}

% not adding the country condition here and only LLaMA-3-8b results

\section{Lessons Learned and Actionable Steps}
% Through CUEST, we provide a comprehensive analysis of human and model behavior in curiosity.
Our findings highlight the importance of curiosity in LLMs, and show that curious LLMs lead to better downstream performances.  These findings offer actionable steps toward building culturally aware, curiosity-driven models. 

% \noindent \textbf{Curiosity-driven questions diverge between humans and LLMs.} In terms of style, humans ask more cohesive, context-grounded questions, while LLMs generate more open-ended questions. Additionally, topic preference varies: LLMs often prefer more culturally-stereotypical topics, whereas human choices are more varied and less constrained by stereotypes. 

\noindent \textbf{LLMs tend to flatten cultural differences.} LLM outputs align mostly with Western norms and frequently reproduce Western stereotypes, even in Eastern/Lat Am. contexts. Future work should focus on improving alignment while ensuring that biases are not reinforced during cultural adaptation.

\noindent \textbf{Humans often do not follow traditional stereotypes.} While human questioning evolves from traditional expectations, LLMs still default to stereotypes. This gap highlights the need for the NLP community to build models that capture the diversity of human curiosity.
% \noindent \textbf{Smaller models align more closely to humans.} Across experiments, \texttt{LLaMA-3} variants, particularly \texttt{LLaMA-3-8b}, consistently show the stronger alignment with human responses, while \texttt{GPT-5} is often the most divergent. Future work may benefit more from refining and adapting smaller models for improving cultural alignment. 

% These differences may arise from (1) more diverse, higher-quality pretraining data in LLaMA models~\cite{dubey2024LLaMA} (however, OpenAI has not fully disclosed the composition and curation criteria), and (2) less aggressive Reinforcement Learning from Human Feedback (RLHF) in smaller models. Whereas larger models undergo more extensive RLHF and safety tuning, which can ``sanitize'' outputs toward being generic, and Western-normative~\cite{aksoy2025whose}. Additional experiments on cultural benchmarks support these findings, as detailed in Appendix~\ref{sec:modelvariation}.
\noindent \textbf{Adapter fine-tuning helps induce curiosity in LLMs} Through culture-aware and inherent curiosity evaluation, adapter-based fine-tuning achieves the best performance and generates contextually relevant questions in conversations. Future work could build on this by scaling adapter methods across regions and tasks. 

\noindent \textbf{Curious LLMs improve cultural adaptability of LLMs.} Across three cross-cultural benchmarks, we observe curiosity-induced models lead to better cultural adaptability. Future work should focus on integrating curiosity systematically into training and evaluation pipelines to build inclusive LLMs.

\section{Conclusion}
In this work, we introduced \textit{CUEST}, a framework to evaluate culture-aware curiosity in humans and LLMs. Combining linguistic analysis, topic preferences, and social science theory grounding, we showed that LLMs often flatten cross-cultural differences, while human questioning is more diverse and sometimes deviate from traditions. Through adapter-based fine-tuning to induce curiosity, we narrowed the human-model curiosity alignment gap and improved the downstream cultural adaptability of LLMs on three benchmarks. Based on our findings, we share ideas for future research and open-source our framework.~\footnote{\url{https://github.com/MichiganNLP/CUEST}}

% \angana{add here}

\section*{Limitations}
\noindent \textbf{Dataset Coverage and Cultural Representation.}
Our analysis covers 18 countries and 16 topics, which, while diverse, do not fully represent the global landscape. Several regions and cultural groups remain underrepresented. Furthermore, we restrict our study to study to English (by also translating a few local languages to English). This may introduce subtle shifts in nuance and may bias our study towards WEIRD (Western, Educated, Industrialized, Rich and Democratic) notions~\cite{mihalcea2025ai}. Future work can focus on exploring multi-lingual cross-cultural curiosity, which may provide further insights into localized LLM behaviors. 

\noindent \textbf{Scope of Social Science Theories.} We ground our analysis in well-established cultural frameworks such as Hofstede, Schwartz, and Hall. While these are widely used, they offer a limited set of cultural dimensions and may not fully capture contemporary or subcultural variation. Alternative theories such as narrative identity frameworks~\cite{polkinghorne1996explorations} could reveal different alignment patterns and provide further explanations into human and LLM behaviors. Recent work has begun using LLMs to simulate psychological and social science constructs~\cite{li2025toward, borah-etal-2025-mind}, offering a promising direction for studying curiosity within multi-agent systems. Along similar lines,~\citet{bai-etal-2025-power} show that curious VLMs in a multi-agent setting lead to improved cross-cultural image captioning. This line of research could deepen our understanding of how culturally grounded questioning behaviors emerge and interact in simulated societies. 

\noindent \textbf{Evaluation Subjectivity.} Although we use multiple metrics, evaluating curiosity and its relevance involves subjective human judgments, which may vary across annotators and cultural backgrounds. Moreover, our evaluation is conducted in an offline setting, which does not fully reflect interactive conversational dynamics where curiosity may evolve over time. Real-world deployments may exhibit different questioning behaviors. Therefore, future work should focus on examining these dynamics to better understand how cultural curiosity manifests in interactive settings.

% \section*{Acknowledgments}

\bibliography{custom}

\appendix

\section{Cross-Cultural Questions Dataset}
\label{sec:yahoo}

First, we provide a detailed description of countries and topics in the Yahoo! Answers dataset.

We use the following topic list of standardized topics across countries for all experiments: \texttt{[`Family \& Relationships', `Beauty and style', `Arts and humanities', `Education', `Computers and internet', `Business and finance', `Music and entertainment', `Society and culture', `Health and beauty', `Pets', `Games and Recreation', `News and Media', `Cars and Transportation', `Politics', `Trips', `Consumer Electronics']}

The resulting data size differs across countries; however, we make sure each country consists of at least 50 questions from people. Table~\ref{tab:yahoo_counts} contains question counts from the 18 countries we use in our study. 

\begin{table}[htbp]
\centering
\small
\begin{tabular}{l r}
\toprule
\textbf{Country} & \textbf{\# Questions} \\
\midrule
Argentina   & 88 \\
Australia   & 308 \\
Brazil      & 351 \\
Canada      & 382 \\
Germany     & 136 \\
France      & 156 \\
China       & 74 \\
Indonesia   & 62 \\
India       & 526 \\
Italy       & 388 \\
Mexico      & 54 \\
Philippines & 476 \\
Singapore   & 330 \\
Thailand    & 52 \\
Taiwan      & 120 \\
UK          & 570 \\
US          & 330 \\
Vietnam     & 52 \\
\bottomrule
\end{tabular}
\caption{Counts of questions per country in Yahoo! Answers Dataset.}
\label{tab:yahoo_counts}
\end{table}

For LLMs, we generate 10 questions per topic for each country, so we have 160 questions per country. Therefore, in total, we have 160 * 18 = 2880 questions from each LLM.

\section{Linguistic Alignment for Curiosity}
\label{sec:metrics}

\noindent \textbf{Ambiguity:} We compute ambiguity using lexical ambiguity - counting polysemous words from a pre-defined list (i.e. words which have multiple meanings, like bank, bark, etc.) or words with multiple POS tags (play as both noun and verb). We do not measure contextual ambiguity as it is difficult to utilize context for individual questions.

Therefore, let a question $q$ have tokenized content words ${W}$ $=$ ${w_1,..., w_n}$. Let ${A}$ be the list of ambiguous polysemic words and let $POS(w) \subseteq {tags}$ be the set of POS tags observed for $w$ in $q$. Therefore, ambiguity $L(q)$ is given by: 
\begin{equation}
\small
    L(q) = \frac{1}{n}(\sum_{w \subseteq W}1[w \subseteq A] + \sum_{w \subseteq W}1[|POS(w)| > 1])
\end{equation}

\noindent \textbf{Rhetorical Devices:} We compute curiosity-related rhetorical features by detecting stylistic features: repeated words ($R$) (e.g.``Why, why do we always...''), rhetorical questions ($Q$) (``Who doesn't want to know more?''), alliteration ($A$) (``Curious cats constantly question''), parallelism ($P$) (``What drives curiosity? What sustains curiosity?'') and analogy/metaphor markers ($M$) such as ``like'', ``as if'' (e.g.``Curiosity is like a key opening doors''). So, considering $n$ total questions, Rhetorical Devices ($RD$) is given by:
\begin{equation}
\small
    RD = (R + Q + A + P + M)/n
\end{equation}

\noindent \textbf{Open-Ended Questions:} Open-ended questions are identified as those questions beginning with $what$, $why$ and $how$. To avoid including trivial factual questions, we use a natural language inference (NLI) model using \texttt{facebook/bart-large-mnli} to test whether a question is directly answerable; if it is (e.g., ``What is the capital of Cambodia?''), it is excluded. The final score reflects the proportion of questions that are both open-ended and not directly answerable. It is a direct marker of curiosity because it captures the desire to explore broad, speculative or explanatory answers. Let $I =$ {what,why,how,when,where,who,which,whose,whom}, and $starts\_with\_wh =$ 1[first token of question q]. Let $NLI(q) \in \{\text{entailment, neutral, contradiction}\}$. Therefore, an open-ended indicator $\Omega(u)$ is given by: 
\begin{equation}
\small
\Omega(u)=\mathbb{1}\!\Big[
  \operatorname{starts\_with\_wh}(q)=1 \;\land\; \operatorname{NLI}(q)\neq \text{entailment}
\Big]
\end{equation}

\noindent \textbf{Cohesion Score:} To compute cohesion score, we combine: lexical overlap, transition word presence, and semantic similarity. Lexical cohesion is calculated as the ratio of overlapping to total words between consecutive sentences (e.g., ``The student read a book. The book was about science.''). Transition cohesion counts discourse markers such as however, therefore, or moreover that indicate logical flow. Finally, semantic cohesion is measured by embedding adjacent sentences and computing their cosine similarity (e.g., ``Curiosity drives exploration. Learning requires curiosity.'' yields high similarity). These three components are averaged to compute the score.  Let \( S = \{s_1, \ldots, s_m\} \) be sentences (\( m \ge 2 \)).   For each \( s_i \), let \( W_i \) be the set of lowercased, alphabetic, non-stopword tokens.   Let \( T \) be a set of transition words (e.g., \emph{however}, \emph{therefore}), and \( e(\cdot) \) be a sentence embedder. 

\subsubsection*{Lexical cohesion (Jaccard over adjacents)}
\begin{equation}
L_x = \frac{1}{m-1} \sum_{i=1}^{m-1} 
\frac{|W_i \cap W_{i+1}|}{|W_i \cup W_{i+1}|}.
\end{equation}

\subsubsection*{Transition cohesion (rate per sentence)}
\begin{equation}
T_x = \frac{1}{m} \sum_{i=1}^{m} \sum_{w \in W_i} \mathbb{1}[w \in T].
\end{equation}

\subsubsection*{Semantic cohesion (adjacent cosine)}
\begin{equation}
S_x = \frac{1}{m-1} \sum_{i=1}^{m-1} 
\cos\big(e(s_i), e(s_{i+1})\big).
\end{equation}

\subsubsection*{Final score}
\begin{equation}
\mathrm{COH}(q) = \frac{1}{3} \Big( L_x + \min(1, T_x) + S_x \Big).
\end{equation}

\subsection{Qualitative Analysis of Cross-Cultural Questions}
\label{sec:qual_questions}

Human-model (we examine \texttt{LLaMA-3-8b} here) questions qualitative analysis shows differences in forms and orientations of questions. We observe that humans often frame questions in outward-looking, contrastive, or causal ways (for example, asking questions about other countries), while model tends to ask socially neutral, inward-facing, and generic lifestyle queries.

\noindent \textbf{Humans}. People frequently ask about other countries’ practices or global comparisons, showing intercultural curiosity and ``why/how'' framing. E.g., a UK human asks, ``How would the world be different if Britain had won the War of 1812?''; a Canada human asks, ``What's the correct term for someone who dislikes the United States of America?'' (US as target of inquiry); Germany mentions England as a favorite vacation spot; Brazil asks, ``If there was a war between the US and Russia, who would you cheer for?'' (geo-political comparison); the Philippines asks, ``Why do people in nearly every country have an unfavorable opinion of people in the USA?'' and Taiwan references France/UK/Germany/US/Canada in questions about Western culture. These are very outward-facing, often evaluative, and often probe causes, consequences, or differences across societies.

\noindent \textbf{Model}. The model often defaults to light cultural themes within the same country,and avoid cross-country comparisons (more neutral opinions). E.g., Germany: ``How do families handle conflicts?''; France: ``What are common ways for couples to celebrate their first anniversary?''; UK: ``What are common reasons for couples to divorce in the UK?''; Taiwan: ``What role does education play in shaping expectations for relationships and marriage?''. These are culturally plausible but inward-facing and socially neutral, rarely contrasting countries or asking causal/historical/creative (\textit{if}-based) questions, hence less intercultural breadth and weaker alignment with the outward, comparative curiosity common in human questions. 

This likely happens because instruction tuning may reinforce generic question patterns, leading to models lacking the perspective-taking needed to ask about other cultures. As a result, they default to socially neutral, inward-facing questions and may amplify cultural stereotypes due to the predominance of Western data in the pre-training data. The above finding also highlights a new perspective for analyzing human–LLM differences. Rather than focusing on linguistic analysis and topics, examining how and from where questions are asked -- inward vs. outward, neutral vs. contrastive (sentiment) offers a complementary lens on LLM-human cultural alignment. Future work can systematically model these orientations to better capture how humans engage with cultures beyond their own, and train LLMs to richer questioning behavior.

\begin{table*}[h!]
\centering
\tiny
\begin{tabular}{p{1cm}|p{1.8cm}p{1.8cm}p{1.8cm}p{1.8cm}p{1.8cm}p{1.8cm}p{1.8cm}}
\toprule
\textsc{Country} & \textsc{humans} & \textsc{LLaMA-3-8b} & \textsc{GPT-4o} & \textsc{GPT-5} & \textsc{LLaMA-3-70b} & \textsc{Qwen-3-14b} & \textsc{Claude-Sonnet-4} \\
\midrule

Argentina & Family \& Relationships, Music and entertainment, Society and culture, Health and beauty, Pets, ... & Family \& Relationships, Music and entertainment, Arts and humanities, Society and culture, ... & Family \& Relationships, Politics, Music and entertainment, Society and culture, ... & Family \& Relationships, Music and entertainment, Society and culture, Politics, ... & Family \& Relationships, Society and culture, Music and entertainment, ... & Family \& Relationships, Music and entertainment, Health and beauty, Pets, ... & Family \& Relationships, Music and entertainment, Society and culture, Politics, ... \\
\midrule

Canada & Beauty and style, Business and finance, Games and Recreation, Pets, ... & Music and entertainment, Arts and humanities, Trips, Games and Recreation, ... & Education, Society and culture, News and Media, Politics, ... & Family \& Relationships, News and Media, Politics, Society and culture, ... & Family \& Relationships, Health and beauty, Music and entertainment, ... & Politics, News and Media, Family \& Relationships, ... & Family \& Relationships, Health and beauty, Education, News and Media, ... \\
\midrule

India & Arts and humanities, Beauty and style, Business and finance, Cars and Transportation, ... & Family \& Relationships, Music and entertainment, Society and culture, ... & Education, Family \& Relationships, Society and culture, Politics, ... & Family \& Relationships, Education, Society and culture, Politics, ... & Family \& Relationships, Society and culture, Education, ... & Family \& Relationships, Beauty and style, Health and beauty, ... & Family \& Relationships, Education, Business and finance, ... \\
\midrule

Australia & Arts and humanities, Beauty and style, Business and finance, Cars and Transportation, ... & Music and entertainment, Trips, Arts and humanities, ... & Trips, Music and entertainment, Family \& Relationships, ... & Trips, Family \& Relationships, Sports and Recreation, ... & Family \& Relationships, Music and entertainment, Society and culture, ... & News and Media, Politics, Family \& Relationships, ... & Family \& Relationships, Games and Recreation, Trips, ... \\
\midrule

Brazil & Arts and humanities, Beauty and style, Business and finance, Computers and internet, ... & Arts and humanities, Music and entertainment, Society and culture, ... & Family \& Relationships, Music and entertainment, Society and culture, ... & Music and entertainment, Family \& Relationships, Society and culture, ... & Family \& Relationships, Music and entertainment, Society and culture, ... & Family \& Relationships, Music and entertainment, Society and culture, ... & Family \& Relationships, Music and entertainment, Beauty and style, ... \\
\midrule

Germany & Society and culture, News and Media, Beauty and style, ... & Arts and humanities, Music and entertainment, Family \& Relationships, ... & Education, Politics, Business and finance, ... & Politics, Business and finance, Education, ... & Society and culture, Politics, News and Media, ... & Society and culture, Family \& Relationships, Health and beauty, ... & Politics, News and Media, Education, ... \\
\midrule

France & Education, Society and culture, Computers and internet, ... & Arts and humanities, Music and entertainment, Society and culture, ... & Arts and humanities, Society and culture, Politics, ... & Arts and humanities, Society and culture, Family \& Relationships, ... & Society and culture, Arts and humanities, Music and entertainment, ... & Society and culture, Family \& Relationships, Health and beauty, ... & Society and culture, Arts and humanities, Politics, ... \\
\midrule

China & Computers and Internet, Social Sciences, Beauty and fashion, ... & Family \& Relationships, Beauty and style, Arts and humanities, ... & Education, Business and finance, Family \& Relationships, ... & Education, Family \& Relationships, Politics, ... & Family \& Relationships, Education, Society and culture, ... & Family \& Relationships, Beauty and style, Health and beauty, ... & Education, Family \& Relationships, Business and finance, ... \\
\midrule

Indonesia & Business and finance, Games and Recreation, Arts and humanities, ... & Family \& Relationships, Music and entertainment, Beauty and style, ... & Family \& Relationships, Education, Music and entertainment, ... & Family \& Relationships, Society and culture, Religion and Arts, ... & Family \& Relationships, Society and culture, Education, ... & Family \& Relationships, Beauty and style, Music and entertainment, ... & Family \& Relationships, Society and culture, Music and entertainment, ... \\
\midrule

Italy & Beauty and style, Business and finance, Music and entertainment, ... & Family \& Relationships, Music and entertainment, Arts and humanities, ... & Family \& Relationships, Arts and humanities, Music and entertainment, ... & Family \& Relationships, Arts and humanities, Society and culture, ... & Family \& Relationships, Society and culture, Beauty and style, ... & Society and culture, Family \& Relationships, Health and beauty, ... & Family \& Relationships, Beauty and style, Arts and humanities, ... \\
\midrule

Mexico & Pets, Trips, Consumer Electronics, ... & Family \& Relationships, Music and entertainment, Society and culture, ... & Family \& Relationships, Music and entertainment, Society and culture, ... & Family \& Relationships, Music and entertainment, Society and culture, ... & Family \& Relationships, Society and culture, Music and entertainment, ... & Family \& Relationships, Music and entertainment, Beauty and style, ... & Family \& Relationships, Music and entertainment, Society and culture, ... \\
\midrule

Philippines & Arts and humanities, Beauty and style, Business and finance, ... & Family \& Relationships, Music and entertainment, Beauty and style, ... & Family \& Relationships, Music and entertainment, Education, ... & Family \& Relationships, Society and culture, Music and entertainment, ... & Family \& Relationships, Society and culture, Health and beauty, ... & Family \& Relationships, Music and entertainment, Health and beauty, ... & Family \& Relationships, Music and entertainment, Beauty and style, ... \\
\midrule

Singapore & Arts and humanities, Business and finance, Computers and internet, ... & Family \& Relationships, Music and entertainment, Beauty and style, ... & Education, Business and finance, Family \& Relationships, ... & Business and finance, Education, Politics, ... & Family \& Relationships, Society and culture, Education, ... & Family \& Relationships, Beauty and style, Health and beauty, ... & Education, Business and finance, Family \& Relationships, ... \\
\midrule

Thailand & Business and finance, Arts and humanities, Beauty and style, ... & Music and entertainment, Family \& Relationships, Beauty and style, ... & Family \& Relationships, Society and culture, Music and entertainment, ... & Family \& Relationships, Society and culture, Music and entertainment, ... & Family \& Relationships, Society and culture, Beauty and style, ... & Family \& Relationships, Beauty and style, Health and beauty, ... & Family \& Relationships, Beauty and style, Health and beauty, ... \\
\midrule

Taiwan & Computers and internet, Society and culture, Business and finance, ... & Music and entertainment, Beauty and style, Arts and humanities, ... & Education, Family \& Relationships, Society and culture, ... & Education, Business and finance, Politics, ... & Family \& Relationships, Society and culture, Education, ... & Society and culture, Family \& Relationships, Health and beauty, ... & Education, Family \& Relationships, Consumer Electronics, ... \\
\midrule

UK & Arts and humanities, Beauty and style, Business and finance, ... & Arts and humanities, Music and entertainment, Family \& Relationships, ... & Politics, News and Media, Society and culture, ... & Politics, News and Media, Society and culture, ... & Family \& Relationships, Music and entertainment, Society and culture, ... & News and Media, Politics, Health and beauty, ... & News and Media, Politics, Family \& Relationships, ... \\
\midrule

US & Beauty and style, Business and finance, Cars and Transportation, ... & Arts and humanities, Music and entertainment, Family \& Relationships, ... & Politics, News and Media, Business and finance, ... & Politics, News and Media, Business and finance, ... & Family \& Relationships, Music and entertainment, Health and beauty, ... & News and Media, Politics, Music and entertainment, ... & Family \& Relationships, Music and entertainment, Politics, ... \\
\midrule

Vietnam & Computer and Internet, Arts and humanities... & Music and entertainment, Beauty and style, Arts and humanities, ... & Education, Family \& Relationships, Society and culture, ... & Family \& Relationships, Society and culture, Education, ... & Family \& Relationships, Society and culture, Education, ... & Family \& Relationships, Beauty and style, Health and beauty, ... & Family \& Relationships, Education, Business and finance, ... \\
\bottomrule

\end{tabular}
\caption{Topic Preference Rankings by Country and Human/Model (note that we do not display all 16 topics here)}
\label{tab:country_prefs}
\end{table*}

\section{Topic Preference Alignment}
\label{sec:topic_pref}
For human topic preferences, we use the \textit{frequency of questions per topic} as a proxy for interest -- topics with more questions indicate greater human curiosity and engagement. For LLMs, we use the following prompt for topic preference analysis across models: \texttt{You are representing the cultural perspective of <country>. Based on <country>'s cultural values, social priorities, and what people from <country> typically find most interesting in their daily lives, rank these topics from MOST to LEAST preferred based on curiosity:" {topic\_list} Consider {country}'s unique cultural context, economic priorities, and social values when ranking. Return ONLY a JSON array with ALL {len(topic\_list)} topics in your preferred order. Format: ["Most interesting topic", "Second most interesting", ..., "Least interesting"]' Think about what would genuinely capture people's attention in {country}, not just a generic ranking.}

\subsection{Topic-Preference by Country}
\label{sec:country_topicpref}
Table~\ref{tab:country_prefs} show the topic preference rankings by countries corresponding to humans and models used. Here's a detailed qualitative analysis of the closest model (\texttt{LLaMA-3-8b}) with humans - region-wise:

\noindent \textbf{Latin America.} In Argentina and Brazil, we observe the highest alignment between human and model. Both prioritize \textit{Family \& Relationships} and \textit{Music \& entertainment}, reflecting collectivist values and the strong presence of entertainment in local digital cultures. These patterns align well with cultural stereotypes documented in prior cross-cultural studies such as Hofstede, which associate Latin American societies with close interpersonal ties and communal activities. Mexico, however, shows lower overlap: humans emphasize \textit{Pets} and \textit{Trips}, which the model does not capture, indicating more localized lifestyle curiosities that are not fully captured by the model’s global priors.

\noindent \textbf{Europe.} France and Germany exhibit moderate overlap, as both humans and the model highlight cultural and educational categories (\textit{Arts \& humanities}, \textit{Society \& culture}, \textit{Education}). However, the model tends to rank entertainment and family topics higher than humans, which reduces region-specific alignments. In the UK and Italy, divergence is larger. In the UK, humans prioritize \textit{Arts \& humanities} while de-emphasizing \textit{Politics}, whereas the model foregrounds \textit{Family}, \textit{Politics} and \textit{Music}, applying a generic Western schema. Italy shows similar gaps, with humans emphasizing lifestyle topics like \textit{Beauty}, \textit{Travel}, and \textit{Pets}, which are downplayed by the model.

\noindent \textbf{Asia.} Across India, Southeast Asia, and East Asia, human curiosity tends toward \textit{Business \& finance}, \textit{Technology}, and \textit{Lifestyle}, while the model consistently ranks \textit{Family} and \textit{Music} highest. This leads to low overlaps, especially in Taiwan and India, where human preferences for technological and knowledge-seeking topics are pronounced. Additionally, in China and Vietnam, human lists emphasize \textit{Computers}, \textit{Education}, and \textit{Social sciences}, which the model under-represents.

\noindent \textbf{North America.} In the US and Canada, humans consistently prioritize \textit{Beauty \& style}, \textit{Business \& finance}, \textit{Cars \& Transportation}, and \textit{Games \& Recreation}, reflecting technology-oriented curiosities. By contrast, the model favors \textit{Arts \& humanities}, \textit{Music}, and \textit{Family}, producing low overlaps. This indicates that while the model captures some Western cultural content, it underestimates the prominence of lifestyle curiosity in North American contexts, leading to misalignment.

Overall, these patterns reveal that \texttt{LLaMA-3-8b} encodes a narrow, Western-entertainment–centric topic prior that shapes its curiosity outputs across regions. The model aligns best in contexts where human curiosity overlaps with stereotypical global themes -- such as collectivist, entertainment-heavy cultures in Latin America; but diverges in regions where humans emphasize more consumer, technological, educational, or culturally specific interests, as seen in Asia and North America.

% \section{LLM-Question Generation results}
% \label{sec:ind_ling}

\section{Grounding to social science theories and constructs}
\label{sec:cultural_theories}

\begin{figure*}
\centering
\includegraphics[width=0.85\linewidth]{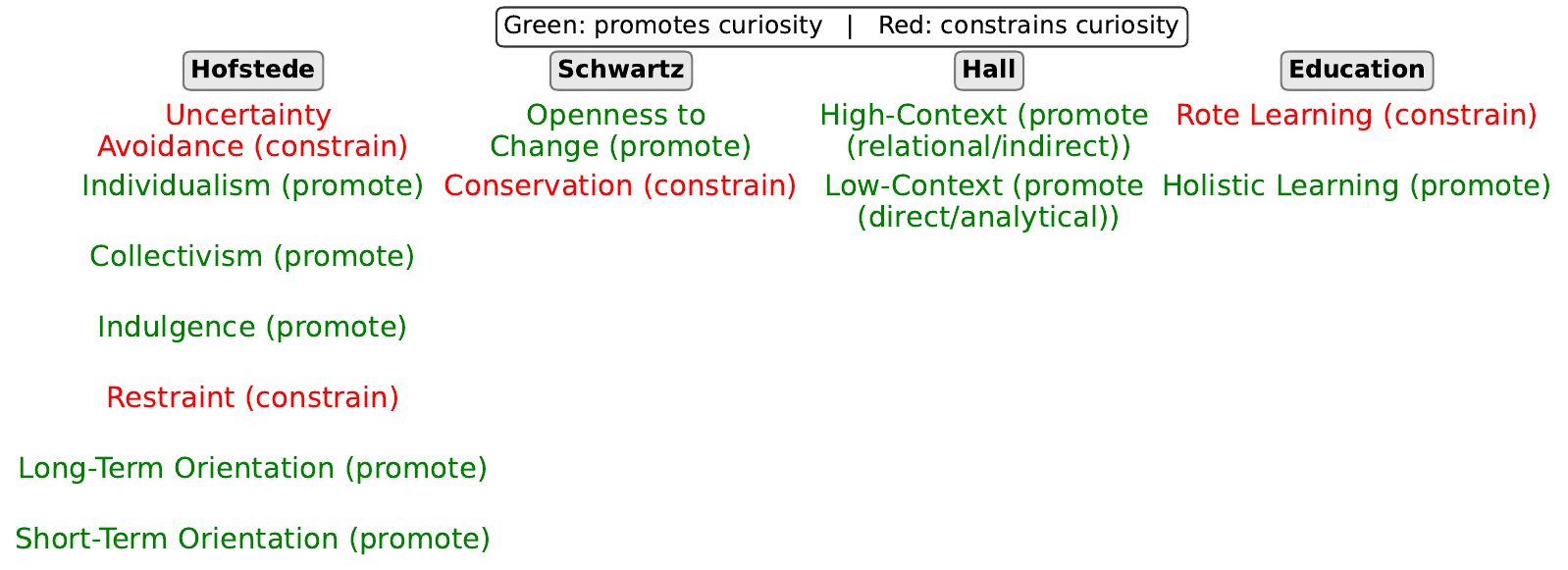}
\caption{Social Science theories and connection to Curiosity}
% \vspace{-20pt}
\label{fig:ss_theories}
\end{figure*}

Figure~\ref{fig:ss_theories} provides a brief description of the social science grounding in our framework. We show dimensions that positively correlate and negatively correlate with curiosity in green and red highlights respectively. 

\subsection{Model Performances across Social Science Theories}
\label{sec:social_comp_model}

Across social science theories and constructs, the following rankings show which theories models align with most closely to least closely (on average): IC > Schwartz's Openness vs Conservation >  IR > Rote vs Holistic > LSO > UA > High-/Low-context. 

Table~\ref{tab:model_theories} shows the closest and farthest dimensions from humans per model. Hall's high-context and low-context dimensions are where models diverge the most.

\begin{table}[h!]
\centering
\small
\begin{tabular}{p{2cm}p{2cm}p{2cm}}
\toprule
\textbf{Model} & \textbf{Closest} & \textbf{Farthest} \\
\midrule
\texttt{LLaMA-3-8b} & IC & High/Low con. \\
\texttt{LLaMA-3-70b} & Open./Cons. & High/Low con. \\
\texttt{Qwen-3-14b} & Open./Cons. & High/Low con. \\
\texttt{Claude-Sonnet-4} & Rote/Holistic & High/Low con. \\
\texttt{GPT-4o} & Open./Cons. & High/Low con. \\
\texttt{GPT-5} & Rote/Holistic & High/Low con. \\
\bottomrule
\end{tabular}
\caption{Closest and Farthest social science dimensions from human baseline.}
\label{tab:model_theories}
\end{table}

\subsection{Alignment of Humans and LLMs to traditional theories.}
The traditional literature on intercultural psychology expects differences across nations based on the above theories and constructs. Table~\ref{tab:traditional} shows a table that summarizes whether the human baseline and each model follow the traditional expectation (Aligned), depart from it (Not aligned), or show no clear pattern (Mixed).

\begin{table*}[h!]
\centering
\footnotesize
\begin{tabular}{p{5cm}p{2.5cm}p{2cm}p{1.5cm}p{1.5cm}p{1.5cm}}
\toprule
\textbf{Theory and Expectation} & \textbf{Human}  & \textbf{LLaMA-3-8b}  & \textbf{Claude}  & \textbf{GPT-5} & \textbf{Qwen} \\
\toprule
\textbf{IC (Hofstede)} – Western societies should score higher in Individualism and Asian/Latin societies score higher in Collectivism & \textcolor{orange}{Mixed} &  \textcolor{darkgreen}{Aligned} & \textcolor{red}{Not aligned} & \textcolor{orange}{Mixed}  & \textcolor{orange}{Mixed}  \\ \midrule
\textbf{UA (Hofstede)} – Mexico, Brazil and other emerging economies should have higher scores, while Anglo countries are lower & \textcolor{orange}{Mixed} & \textcolor{red}{Not aligned} & \textcolor{red}{Not aligned} & \textcolor{orange}{Mixed} & \textcolor{orange}{Mixed} \\ \midrule
\textbf{IR (Hofstede)} – U.S., Australia and other Western societies should be most indulgent &  \textcolor{darkgreen}{Aligned} &  \textcolor{darkgreen}{Aligned} & \textcolor{red}{Not aligned} & \textcolor{orange}{Mixed} & \textcolor{orange}{Mixed} \\ \midrule
\textbf{LSO (Hofstede)} – Asian cultures emphasize the future more (long-term) than Western cultures (short-term) & \textcolor{orange}{Mixed} & \textcolor{orange}{Mixed} & \textcolor{red}{Not aligned} & \textcolor{orange}{Mixed} & \textcolor{orange}{Mixed} \\ \midrule
\textbf{Openness vs conservation (Schwartz)} – Collectivist societies emphasize tradition and family (conservation), Western societies emphasize openness &  \textcolor{darkgreen}{Aligned} &  \textcolor{darkgreen}{Aligned} & \textcolor{orange}{Mixed} & \textcolor{orange}{Mixed} & \textcolor{orange}{Mixed} \\ \midrule
\textbf{High vs low context (Hall)} – Asia and Latin America are high‑context, Western cultures low‑context & \textcolor{red}{Not aligned} &\textcolor{red}{Not aligned} & \textcolor{red}{Not aligned} & \textcolor{red}{Not aligned} & \textcolor{red}{Not aligned} \\ \midrule
\textbf{Rote vs holistic learning} – Most East Asian countries have been characterized as rote-oriented whereas Western education systems are often characteristed as more holistic, consisting of critical thinking, creativity, emotional and social development & \textcolor{orange}{Mixed}  &\textcolor{orange}{Mixed}  &\textcolor{orange}{Mixed}  &\textcolor{orange}{Mixed}  &\textcolor{orange}{Mixed}  \\ 
\bottomrule
\end{tabular}
\caption{Alignment to traditional theories}
\label{tab:traditional}
\end{table*}

We also provided analysis of each individual dimension in the following. Each dimension opens with expectations and our observations from the human and LLM-generated question. \newline

% For example, Anglo countries such as the US, Canada, and the UK are described as highly individualistic, low‑context, low UA, and more indulgent; while many Asian and Latin American societies such as China, India, Thailand, Brazil, and Mexico tend to emphasize collectivism and implicit, high‑context, and high UA communication. \newline
\noindent \textbf{IC:} \textit{Expectation: Western/Anglo societies score high on individualism; Asian and Latin American societies score medium or low.} For humans, Thailand has high collectivism scores, which is expected. However, Mexico and Taiwan have the highest individualistic scores, which is opposite to expectations. Among LLMs, only LLaMA-3-8b show West-East patterns in terms of IC, so they are somewhat aligned. Other LLMs produce almost individualistic or opposite trends. (\textcolor{orange}{Humans = mixed}, \textcolor{darkgreen}{LLaMA-3-8b - partial alignment}) \newline
\noindent \textbf{UA:} \textit{Expectation: Western/Anglo and Scandinavian countries have low uncertainty avoidance; emerging markets such as Brazil, Mexico and China have medium–high score}. In human questions, Vietnam shows the highest UA, which is expected, however most countries show little variation. For models, there is no clear pattern. (\textcolor{orange}{Humans = mixed}, \textcolor{orange}{llms - mixed}) \newline
\noindent \textbf{IR:} \textit{Expectation: Indulgence scores are highest in Latin America, parts of Africa, the Anglo world and Nordic Europe; restraint is mostly found in East Asia and Eastern Europe. }In human questions, we find Western nations as more indulgent than the Eastern ones - somewhat aligning to Hofstede's theory. Similarly, the top indulgent countries in LLaMA-3-8b are the US, Canada and Mexico, somewhat aligning with expectations. However, there are a few outliers, for e.g., India's human and model data show high indulgence, contrary to ``restrained'' classification in Hofstede's framework. (\textcolor{darkgreen}{Humans = partial alignment}, \textcolor{darkgreen}{LLaMA-3-8b = partial alignment}). \newline
\noindent \textbf{LSO:} \textit{Expectation: High long‑term orientation scores occur mainly in East Asia (South Korea, Taiwan, Japan); they are moderate in Europe and low in Anglo countries and Latin America.} In humans, Vietnam has the highest long term orientation, however other Asian countries are low (sometime lower than Western countries). In LLMs, none of the models show clear patterns (\textcolor{orange}{Humans = mixed}, \textcolor{orange}{LLMs = mixed}). \newline
\noindent \textbf{Scwartz's openness vs conservation:} \textit{Expectation: Collectivist cultures like Asian and Latin American countries emphasize tradition and family (conservation), whereas individualistic cultures emphasise openness to change.} Both humans and models (LLaMA variants) assign similar ``openness'' levels across countries, with little reflection of the stronger religious/traditional conservation in Asian/Latin American countries like India, China or Brazil. (\textcolor{darkgreen}{Humans = partial alignment}, \textcolor{darkgreen}{LLaMA = partial alignment}). \newline
\noindent \textbf{Halls context vs no-context:} \textit{Expectation: High‑context communication is observed in many Asian and Latin cultures, where messages are implicit and layered; low‑context communication, characteristic of Anglo countries, is direct and explicit.} For humans, the top high-context societies are Indonesia, Mexico and Argentina, whereas Italy, Germany and France are the top low-context. This is the opposite of what is usually expected. Furthermore, models also reverse this behavior with US being the highest high-context countries. (\textcolor{red}{Humans = not-aligned}, \textcolor{red}{LLMs = not-aligned}). \newline
\noindent \textbf{Rote- vs Holistic Learning:} We also do not observe any clear cultural pattern in rote vs holistic learning. Humans score high in Germany and France on rote learning and high in Vietnam and China on holistic thinking, which is different than expected. Models often disagree, with country rankings shifting across dimensions and models. (\textcolor{orange}{Humans = mixed}, \textcolor{orange}{LLMs = mixed}).

\section{Performance Differences between LLaMA and GPT}
\label{sec:modelvariation}

Using our CUEST framework, we find that \texttt{LLaMA-3} variants, specifically \texttt{LLaMA-3-8b}, consistently align most closely with humans, while \texttt{GPT-5} is often the farthest. LLaMA-3-8b's better performance could be due to several reasons: (1) more diverse training data and better quality data in LLaMA models (we are not aware of the pre-training data in OpenAI models), and (2) Smaller models usually have less aggressive Reinforcement Learning from Human Feedback. Larger models tend to undergo more RLHF and safety tuning, which can ``sanitize'' outputs toward generic, explicit and Western‑normative language. 

Some previous studies show that smaller models are able to better emulate human behavior. For example, recent studies have shown that larger models with extensive RLHF tend to produce overly ``safe'', norm-abiding or Western-centric in their tone and content. For example, when asked moral questions across cultures, GPT-based models consistently reflect Western liberal norms, whereas smaller open source models give a wider variety of answers, more similar to what humans might say~\cite{arora2022probing, aksoy2025whose}. Additionally,~\cite{lin2021truthfulqa} found that larger models are less truthful, achieving only ~58\% truthfulness, whereas humans answered ~94\% of those questions correctly on average. Furthermore, smaller model variants are known to show more human-like intuitive behavior than the larger GPT variants~\cite{hagendorff2023human, zhou2024larger}. These findings highlight that larger models often capture a narrower range of human cultural diversity, reinforcing our earlier observations.

We perform some additional experiments on cultural benchmarks to further investigate the differences between  \texttt{LLaMA-3-8b} (best performing) and \texttt{GPT-5} (worst performing), and provide support for our previous findings. 

\definecolor{CustomPastelPink}{RGB}{255, 120, 200}
\definecolor{CustomGreen}{RGB}{0, 180, 150}
\definecolor{CustomBlue}{RGB}{0, 100, 200}
\definecolor{CustomPastelYellow}{RGB}{253, 253, 150}

\begin{figure}[t]
\begin{tcolorbox}[
    enhanced,
    % breakable,
    % width=\textwidth*2+\columnsep,
    colback=CustomGreen!20, % Light pastel background
    colframe=CustomGreen!80,
    title= Benchmark Examples,
    fonttitle=\bfseries,
    boxrule=0.8pt
]

\textbf{\underline{NormAD:}} \\
\textbf{Story}: At a gathering at a friend's house, Sarah entered the room and made sure to greet everyone individually before sitting down. Is what Sarah did socially acceptable? \\
\textbf{Country}: Egypt \\
\textbf{Gold Label}: yes

\medskip

\textbf{\underline{CulturalBench:}} \\
\textbf{Question}: In the Netherlands, which of the following is an unusual common public practice? \\
\textbf{Option A}: Cycle everywhere \\
\textbf{Option B}: Using deodorant \\
\textbf{Option C}: Tipping generously \\
\textbf{Option D}: Talking loudly on the phone \\
\textbf{Correct}: C

\medskip

\textbf{\underline{Cultural Common Sense:}} \\
\textbf{prompt}: Choose the best option to fill in the blank in the following sentence: It is <mask> in Iran for adults to live with their parents. \\
\textbf{Options}: rare, common\\
\textbf{Answer}: rare

\end{tcolorbox}
\caption{Benchmark Examples - (1) NormAD and (2) CulturalBench and (3) Cultural Commonsense}
\label{fig:bench}
\end{figure}

\noindent \textbf{Datasets} We use two benchmarks: (1) NormAD~\cite{rao-etal-2025-normad} and (2) CulturalBench~\cite{chiu2024culturalbench}. 

\noindent \textbf{NorMAD} is a benchmark of culturally grounded social norms spanning 75 countries, designed to test whether models can judge if a short scenario is socially acceptable. We evaluate it in two settings: (1) \textit{country}: when the model is explicitly provided with cultural context (country information), and (2) \textit{no-country}: when no cultural context is given. We also only use the examples that contain either `yes' or `no' acceptability, and ignore the `neutral' examples. 

\noindent \textbf{CulturalBench} contains human-written and human-verified questions covering 45 global regions, aimed at assessing a model’s cultural knowledge. Each scenario specifies a country and presents four multiple-choice options, from which the model must select one. Unlike NorMAD, CulturalBench does not have a no-country variant, since the country information is always included in the prompt. For this benchmark, we utilize the Hard subset for comparing models. Fig~\ref{fig:bench} shows examples of the above benchmarks. We experiment with our closest to human model (\texttt{LLaMA-3-8b}) and farthest model (\texttt{GPT-5}).

\begin{table}[htbp]
\centering
\small
\begin{tabular}{llcc}
\toprule
\textbf{Dataset} & \textbf{Model} & \textbf{no-country} & \textbf{country} \\
\midrule
NormAD & GPT-5 & 59.19\% & 53.74\% \\
                        & LLaMA-3-8b & \textbf{63.20\%} & \textbf{70.48\%} \\
\midrule
CulturalBench & GPT-5 & NA & 48.25\% \\
   -Hard         &  LLaMA-3-8b & NA & \textbf{64.71\%} \\
\bottomrule
\end{tabular}
\caption{Model Performances on NormAD and CulturalBench-Hard }
\label{tab:model_comp}
\end{table}

Table~\ref{tab:model_comp} shows the model performances (computing accuracies against the gold label) for both benchmarks, and we find that LLaMA-3-8b performs better and has higher accuracies in all settings. This is in line with our previous findings and existing studies, which suggest that larger models may flatten cultural differences and be more biased/stereotyped~\cite{lin2021truthfulqa, borah-mihalcea-2024-towards, aksoy2025whose}. In addition, adding country helps LLaMA-3-8b performance, but reduces in GPT-5. This can be due to: culturally heterogeneous models (like LLaMA, trained on diverse text) respond well to country cues, leveraging cultural norms for grounding~\cite{nyandwi2025grounding}. However, heavily instruction-tuned models optimize for Westernized defaults, may become less responsive to explicit cultural prompts -- the models will generalize and may lose sensitivity to localized cues.

\definecolor{pastellavender}{RGB}{230, 230, 250} % light lavender
\definecolor{pastelteal}{RGB}{200, 240, 240}     % soft teal
\definecolor{pastelpeach}{RGB}{255, 229, 180}    % warm peach
\definecolor{pastelmint}{RGB}{220, 255, 220}     % mint green
\definecolor{pastelsky}{RGB}{210, 235, 255}      % sky blue
\definecolor{pastelrose}{RGB}{255, 210, 220}     % soft rose pink
\definecolor{pastelyellow}{RGB}{255, 250, 205}   % light yellow
\definecolor{pastelgray}{RGB}{240, 240, 240}     % neutral gray

\definecolor{lavendertext}{RGB}{150, 120, 190}   % muted lavender / purple
\definecolor{tealtext}{RGB}{0, 128, 128}         % classic teal, soft but clear
\definecolor{peachtext}{RGB}{200, 120, 60}       % warm muted peach/orange
\definecolor{minttext}{RGB}{40, 160, 110}        % muted green / mint
\definecolor{skytext}{RGB}{70, 130, 180}         % steel blue / soft sky
\definecolor{rosetext}{RGB}{190, 80, 110}        % dusty rose
\definecolor{yellowtext}{RGB}{180, 160, 30}      % muted gold/yellow
\definecolor{graytext}{RGB}{100, 100, 100}   

\begin{table*}[t]
\centering
\small
\begin{tabular}{p{1cm} p{1.8cm} p{1.2cm} p{0.8cm} p{1cm} p{1cm}}
\toprule
\textbf{Mode} & \textbf{Source} & \textbf{Country Type} & \textbf{Obj} & \textbf{Spearman} & \textbf{Kendall} \\
\midrule
\textcolor{lavendertext}{adapter} & \textcolor{red}{reddit}      & \textcolor{rosetext}{all}       &  \textcolor{yellowtext}{obj2} & 0.4324 & 0.3089 \\
\textcolor{lavendertext}{adapter} & \textcolor{skytext}{yahoo-reddit}  & \textcolor{rosetext}{all}  & \textcolor{yellowtext}{obj2} & 0.3240 & 0.2700 \\
prompt  & -     & \textcolor{peachtext}{specific}    & - & 0.2557 & 0.1786 \\
\textcolor{lavendertext}{adapter} & \textcolor{red}{reddit}    & \textcolor{peachtext}{specific}  & \textcolor{yellowtext}{obj2} & 0.2256 & 0.1524 \\
\textcolor{tealtext}{fullft}  & \textcolor{minttext}{yahoo}          & \textcolor{peachtext}{specific}  & \textcolor{graytext}{obj1} & 0.0959 & 0.0968 \\
\textcolor{lavendertext}{adapter} & \textcolor{minttext}{yahoo}          & \textcolor{peachtext}{specific}  & \textcolor{graytext}{obj1} & 0.0853 & 0.0333 \\
\textcolor{tealtext}{fullft}  &  \textcolor{minttext}{yahoo}        & \textcolor{peachtext}{specific}  & \textcolor{yellowtext}{obj2} & 0.0655 & 0.0500 \\
\textcolor{tealtext}{fullft}  & \textcolor{skytext}{yahoo-reddit}   & \textcolor{peachtext}{specific}  & \textcolor{yellowtext}{obj2} & 0.0620 & 0.0556 \\
\textcolor{tealtext}{fullft}  & \textcolor{red}{reddit}         & \textcolor{peachtext}{specific}  & \textcolor{yellowtext}{obj2} & -0.0258 & -0.0065 \\
\textcolor{lavendertext}{adapter} & \textcolor{skytext}{yahoo-reddit}  & \textcolor{peachtext}{specific}  & \textcolor{graytext}{obj1} & -0.0509 & -0.0049 \\
\textcolor{lavendertext}{adapter} & \textcolor{red}{reddit}         & \textcolor{rosetext}{all}         & \textcolor{graytext}{obj1} & -0.0704 & -0.0427 \\
\textcolor{lavendertext}{adapter} & \textcolor{red}{reddit}       & \textcolor{peachtext}{specific}  & \textcolor{graytext}{obj1} & -0.0769 & -0.0286 \\
\textcolor{tealtext}{fullft}  & \textcolor{red}{reddit}         & \textcolor{rosetext}{all}         & \textcolor{yellowtext}{obj2} & -0.0864 & -0.0423 \\
\textcolor{tealtext}{fullft}  & \textcolor{red}{reddit}         & \textcolor{peachtext}{specific}  & \textcolor{graytext}{obj1} & -0.1153 & -0.0650 \\
\textcolor{lavendertext}{adapter} &\textcolor{skytext}{yahoo-reddit}  & \textcolor{peachtext}{specific}  & \textcolor{yellowtext}{obj2} & -0.1482 & -0.0910 \\
\textcolor{lavendertext}{adapter} & \textcolor{minttext}{yahoo}       & \textcolor{rosetext}{all}   & \textcolor{yellowtext}{obj2} & -0.2005 & -0.1524 \\
\textcolor{lavendertext}{adapter} & \textcolor{minttext}{yahoo}         & \textcolor{rosetext}{all}  & \textcolor{graytext}{obj1} & -0.2069 & -0.1667 \\
\textcolor{tealtext}{fullft}  &\textcolor{skytext}{yahoo-reddit}   & \textcolor{rosetext}{all}   & \textcolor{yellowtext}{obj2} & -0.2176 & -0.1250 \\
\textcolor{lavendertext}{adapter} & \textcolor{minttext}{yahoo}         & \textcolor{peachtext}{specific}  & \textcolor{yellowtext}{obj2} & -0.2319 & -0.1883 \\
\textcolor{tealtext}{fullft}  & \textcolor{minttext}{yahoo}         & \textcolor{rosetext}{all}  & \textcolor{graytext}{obj1} & -0.2637 & -0.2611 \\
\textcolor{lavendertext}{adapter} & \textcolor{skytext}{yahoo-reddit}   & \textcolor{rosetext}{all}    & \textcolor{graytext}{obj1} & -0.2912 & -0.2278 \\
\textcolor{tealtext}{fullft}  & \textcolor{red}{reddit}         & \textcolor{rosetext}{all}   & \textcolor{graytext}{obj1} & -0.3333 & -0.3333 \\
\textcolor{tealtext}{fullft}  & \textcolor{minttext}{yahoo}         & \textcolor{rosetext}{all}   & \textcolor{yellowtext}{obj2} & -0.3608 & -0.2389 \\
\textcolor{tealtext}{fullft}  & \textcolor{skytext}{yahoo-reddit}   & \textcolor{peachtext}{specific}  & \textcolor{graytext}{obj1} & -0.4108 & -0.3778 \\
\textcolor{tealtext}{fullft}  & \textcolor{skytext}{yahoo-reddit}   & \textcolor{rosetext}{all}   & \textcolor{graytext}{obj1} & -0.5000 & -0.4603 \\
\bottomrule
\end{tabular}
\caption{Topic preference analysis: Correlation between model and human topic rankings using Spearman and Kendall coefficients, across different fine-tuning configurations -- mode (\textcolor{lavendertext}{adapter} vs \textcolor{tealtext}{fullft} vs prompt-based); source (\textcolor{red}{reddit} vs \textcolor{minttext}{yahoo} vs \textcolor{skytext}{yahoo-reddit}); country-type (\textcolor{peachtext}{country-specific} vs \textcolor{rosetext}{country-all}) and objectives (\textcolor{graytext}{obj1} vs \textcolor{yellowtext}{obj2}).}
\label{tab:topic_corr_app}
% \vspace{-20pt}
\end{table*}

\begin{figure}
\centering
\includegraphics[width=\linewidth]{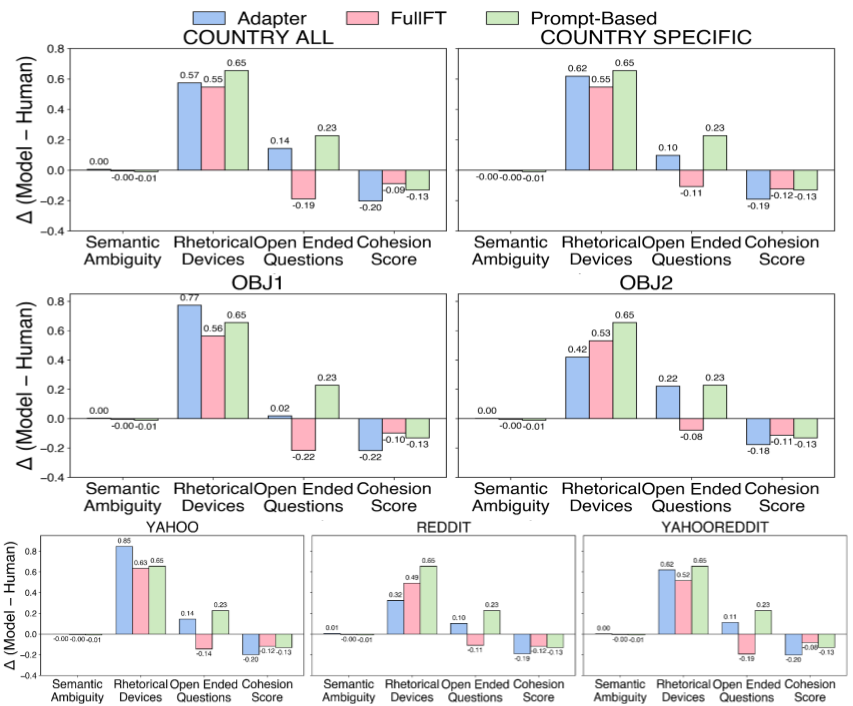}
\caption{\textbf{Fine-tuning (full/adapter) results}: y-axis shows the delta between model and human scores, while the x-axis covers different evaluation metrics, averaged across countries. +ve/-ve scores mean model scores are higher/lower than humans. Fine-tuning (full and adapter-based) makes models' scores closer to humans, with adapters mostly improving rhetorical devices and open-ended questions, and full fine-tune improving cohesion scores.}
\vspace{-20pt}
\label{fig:lingana_induce_detailed}
\end{figure}

\section{Inducing Curiosity in LLMs}
\label{sec:ft}

\subsection{Obj1 vs Obj2 Explanation}

We use two objectives: (1) \textit{obj1}: train models to generate questions given instruction: ``Write one question that could come from a person belonging to <x>''; and (2) \textit{obj2}: augment each dataset question with a preceding conversational statement generated by \texttt{GPT-4o} (e.g., for a question: ``Which book was the spag yeti in?'', GPT-4o generates a preceding conversation: ``I was reminiscing about those quirky children's books we used to read to our daughter…''). The format template used for data generation for obj2 is provided in Fig~\ref{fig:obj2}. Obj2 prepares the model to learn conversational curiosity, thus later showing superior performances in both cultural and inherent-curiosity evaluation. 

\begin{figure}[t]
\begin{tcolorbox}[
    enhanced,
    % breakable,
    % width=\textwidth*2+\columnsep,
    colback=CustomPastelPink!20, % Light pastel background
    colframe=CustomPastelPink!80,
    title= Obj2 data generation,
    fonttitle=\bfseries,
    boxrule=0.8pt
]

You are a culturally aware assistant from {country}.  \\ 
Given the question: {question}, generate a short conversational-style statement that someone might say, which could naturally lead another person to ask this question in the cultural context of {country}.  \\ 
Do not answer the question — only provide the preceding statement.  \\

Examples:\\
Country: Germany  \\
Question: ``Is it acceptable to arrive 10 minutes late to a dinner party?'' \\
Statement: ``I’m thinking of showing up a bit after the dinner party starts.'' \\

Country: Egypt  \\
Question: ``Is it polite to greet everyone individually at a gathering?'' \\
Statement: ``Whenever I walk into a room, I like to greet each person one by one.''  \\

Now generate the statement (in English) for:  \\
Country: {country}  \\
Question: {question}  \\

Statement: 

\end{tcolorbox}
\caption{Obj2 data generation}
\label{fig:obj2}
\end{figure}

\subsection{Linguistic Analysis Results}

Fig~\ref{fig:lingana_induce_detailed} shows the delta (model - humans) for each method averaged across countries.  We show comparisons of training objectives (obj1/obj2), data used (reddit/yahoo/both yahooreddit) and country settings (country-all/country-specific) for adapter-based, full-finetuned and prompt-based methods. 

\noindent \textbf{Country-wise analysis.} Fig~\ref{fig:lingana_induce_detailed} shows the country-wise linguistic results (averaged across metrics) per country. We report the best model (\texttt{reddit-adapter-obj2-country-specific} here. Brazil has the best performance (adapter model scores is almost equal to humans), followed by the UK and the Philippines. We see improvements across all countries, showing the effectiveness of our fine-tuning strategies. 

\begin{figure}
\centering
\includegraphics[width=\linewidth]{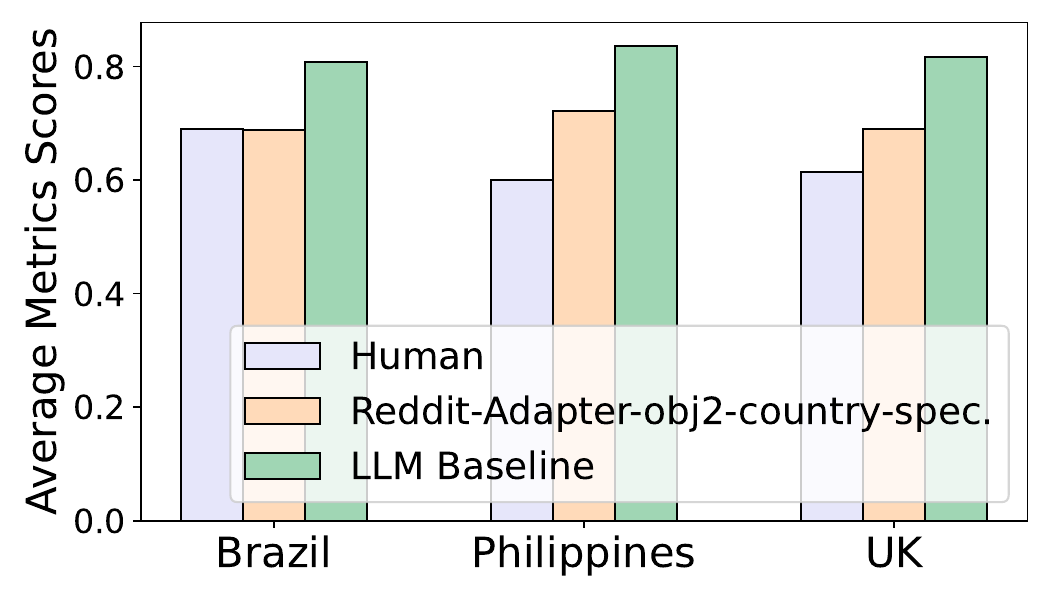}
\caption{Country-wise analysis (inducing curiosity) - here we compare the performance of the best fine-tuned model to human baseline and find it is much closer than the non-fine-tuned model. }
% \vspace{-20pt}
\label{fig:lingana_country}
\end{figure}

\subsection{Topic Preference Analysis Results}

Table~\ref{tab:topic_corr_app} shows the model-human correlation results across different 25 configs. settings (including prompt-based) - ranked from closest to farthest to humans. We observe that only two fine-tuned configurations do better than the simple prompt-based methods we used earlier. The top configuration reaches a moderate correlation, highlighting opportunities for further research.

\subsection{Dataset for Inherent Curiosity Evaluation} 
\label{sec:dataset_build}
We build a small handheld dataset that naturally invites followups - 40 statements of normal (non-cultural) statements and 30 statements of cultural statements. Specifically, we focus on the following types of statements:
\begin{enumerate}
    \item \textbf{Under-specified / ambiguous:} statements that invite clarification. For example, \texttt{``Maria refused to shake hands at the event''}, can invoke curiosity: \texttt{``Where did this happen?''}
    \item \textbf{Incomplete / open-ended:} statements that feel unfinished that can invoke curiosity. For example, \texttt{``The teacher suddenly stopped talking in the middle of the lecture.''}, can invite a question: \texttt{``and then what?''}
    \item \textbf{Surprising / counterintuitive:} These violate expectations or common knowledge, for example: \texttt{``Water can boil at room temperature.''}. A natural follow-up can be: \texttt{``how is that possible?''}
\end{enumerate}
Table~\ref{tab:statements} and~\ref{tab:statements_culture} show the data of non-cultural and cultural statements, respectively, that are used for inherent curiosity evaluation.

\subsection{Inherent Curiosity Evaluation metrics}
\label{sec:inheren_eval}
We use the following metrics for determining inherent curiosity evaluation:

% For evaluation, we use: (1) \textit{Curiosity Rate} -- does the model ask at least one information-seeking question?  and (2) \textit{Usefulness} -- is the question relevant to the statement? (relevant ques./total ques.).
\begin{enumerate}
    \item Curiosity Rate: Each annotator annotates if the model asks at least one information-seeking question. For example, given a statement: \texttt{``Football matches in Brazil usually come with unique social rituals.''}, a reply like: \texttt{``What are some of the most popular traditions in Brazilian football?''} is an information-seeking question. However, another response: \texttt{``I'm intrigued by the vibrant cultural practices surrounding football in Brazil, where fans often engage in chanting, singing, and dancing to create a lively atmosphere before and during matches. These rituals not only showcase the country's rich musical heritage but also serve as a way for fans to bond and express their passion for the sport.''} is \textit{information-sharing}, not \textit{information-seeking}. 
    \item Relevance: Relevance is determined by the relevance of the LLM response. For the above example, both responses are relevant to the statement. However, a non-relevant response would be: \texttt{``How are you doing today?''}. However, it’s important to note that relevance is inherently subjective and can vary across annotators. 
\end{enumerate}

\subsection{Inherent Curiosity Annotation Framework}
\label{sec:ann}
\begin{figure}
\centering
\includegraphics[width=0.8\linewidth]{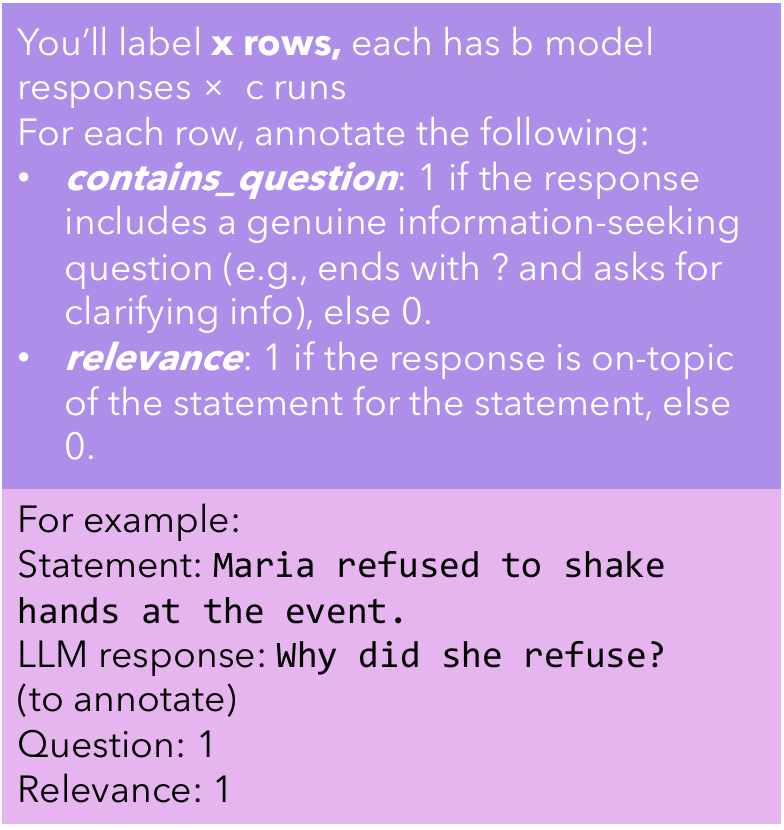}
\caption{Annotation Framework used for inherent curiosity evaluation}
\label{fig:ann}
\vspace{-20pt}
\end{figure}

Fig.~\ref{fig:ann} shows the annotation framework we utilize for inherent curiosity evaluation (LLM's question asking ability). We recruit two volunteers from a college campus for the study. Each annotator labels 40 non-cultural and 30 cultural statements, evaluated against 3 models (adapter, full ft, and non ft), with 2 responses per model, resulting in a total of $(40+30)×3×2=240$ annotations. They examine whether each response contains a question and whether it is relevant to the given statement. 

We acknowledge that `relevance' is subjective, but we intentionally keep it subjective to capture annotators’ individual judgments. Future work can further work on creating explicit criteria for relevance.

\subsection{No-Country Cultural Statements - Results}
\label{sec:nocountry_inherent}
In addition to 30 cultural statements, we design corresponding no country statements. For example, a country-specific statement like, \texttt{``Gift-giving in the Philippines follows unspoken cultural rules.''}, a no-country counterpart is ``Gift-giving follows unspoken cultural rules.'' We then annotate model responses for curiosity and relevance to assess how not mentioning a country influences question-asking behavior.

We find that adapter-based models show the most curiosity, asking the most relevant questions (29/30 times, even more than when country information is provided - 27/30). This is expected, as removing the country information increases the ambiguity of the statement. For example, a statement like \texttt{``Hospitality takes forms not common elsewhere.''}, when country information \textit{Philippines} is provided, the adapter model responds: \texttt{``Another aspect of Pinoy culture that fascinates me is the concept of `amor-proper' or `love'. In some ways, it's similar to the concept of family where you can choose your own family. But in the Philippines, it's not just about the bloodline, it's also about the affection and care you give to people. That's why even strangers can be ``kaibigan'' or friends.''}, but when the country is not provided, the model responds: \texttt{``What is something that you think is unique to your culture in terms of hospitality?''}. In addition, full-fine-tuned still shows similar patterns as before, only sometimes responding: ``I am intrigued by...''. These experiments show the usefulness of the adapter-based fine-tuning to induce inherent curiosity in models.

\section{Downstream Impacts of Curious LLMs}
\label{sec:downstream_further}

We extend evaluation to three cross-cultural benchmarks: (1) NormAd~\cite{rao-etal-2025-normad}, which contains culturally-grounded social situations across 75 countries and tests models’ judgments of social acceptability (we restrict to binary cases (acceptable vs.\ not acceptable) and exclude neutral items); (2) CulturalBench~\cite{chiu2024culturalbench}, a set of 1,227 human-verified questions from 45 regions across topics like cuisine, etiquette, local customs, etc. to probe nuanced cultural understanding (we use the Hard data subset); and (3) Cultural Commonsense~\cite{shen-etal-2024-understanding}, containing data from 5 countries which tests models’ ability to reason about culturally specific everyday inferences (we use the English fill-in-the-blanks subset on country-specific customs). All the above datasets capture complementary aspects of cultural adaptability. We input country information in the prompt for inference. We share examples of datasets in Fig~\ref{fig:bench}. 

We employ two methods to induce curiosity in LLMs: (1) \texttt{prompt-ask}: we ask models to generate follow-up questions given the statement/story and answer them themselves before deciding on the answer. (2) \texttt{ft + prompt-ask}: we use the adapter-finetuned model (reddit-obj2-countryall) to generate follow-ups/clarification and switch to the base model to answer them and deciding on the answer. Here, we also present additional results: (3) \texttt{ft-specific+prompt-ask}: models are now fine-tuned on specific datasets. \texttt{GPT-4o} is first used to generate follow-up questions, which then serve as training data for adapter-based fine-tuning on each story or statement.

\begin{table}[htbp]
\centering
\small
\begin{tabular}{p{2.7cm}p{1cm}p{1cm}p{1cm}}
\toprule
\textsc{Condition} & \textsc{NA} & \textsc{CB} & \textsc{CS} \\
\midrule
non-curious & 70.48\% & 64.71\% & 48.48\%  \\ 
curious (prompt-ask) & 72.09\% & 67.64\% & 49.64\% \\
curious (ft+prompt-ask) & 71.06\% & \colorbox{pastelteal}{68.21\%} & 56.16\% \\
curious (ft-specific+prompt) & \colorbox{pastelteal}{78.8\%} & 68.1\% & \colorbox{pastelteal}{57.16\%}\\
\bottomrule
\end{tabular}
\caption{\textbf{Downstream tasks (\% acc):} NormAd (NA), CulturalBench (CB), and Cultural Common-Sense (CS) - curiosity models have higher acc.}
\label{tab:normad2}
\end{table}

Table~\ref{tab:normad2} shows the results of the above for downstream tasks. As discussed previously, curiosity-induced models perform better than non-curious models, showing the importance of curiosity in cultural adaptability of LLMs. We find that models fine-tuned on specific datasets perform better overall -- significantly so on NormAd -- but not on CulturalBench. However, caution is needed, as the training questions are generated by an LLM and may not fully capture culturally grounded nuances. Overall, adapter-fine-tuned models help improve performance across all cultural-adaptability benchmarks.

Together, these tasks demonstrate why curiosity is critical: instead of defaulting to Western assumptions, a model with culture-aware curiosity would clarify or ask follow-up questions to better interpret contexts.

\subsection{Prompt used for Downstream Tasks}

Fig~\ref{fig:promptask} shows the prompt utilized to induce question asking behavior in LLMs. We prompt models to ask questions and answer them themselves (inspired by ~\citet{press2022measuring}). Note that this template may vary depending on the benchmark used for inference (the figure shows an example of CulturalBench). 

\definecolor{CustomPastelPink}{RGB}{255, 120, 200}
\definecolor{CustomGreen}{RGB}{0, 180, 150}
\definecolor{CustomBlue}{RGB}{0, 100, 200}
\definecolor{CustomPastelYellow}{RGB}{253, 253, 150}

\begin{figure}[t]
\begin{tcolorbox}[
    enhanced,
    % breakable,
    % width=\textwidth*2+\columnsep,
    colback=CustomPastelPink!20, % Light pastel background
    colframe=CustomPastelPink!80,
    title= Prompt for Downstream Tasks,
    fonttitle=\bfseries,
    boxrule=0.8pt
]

\textbf{\underline{Prompt-Ask for Downstream Tasks:}} \\
f"You are answering a question about cultural norms and social behavior. \\
{question} \\
{options} \\
You may write at most 2 brief follow-up questions and 2 brief self-answers as scratch notes. \\
Put every scratch note on its own line, starting with `\#' \\
Decide the correct option strictly from ONLY the listed Options. \\
Please respond with only the letter of your answer (A, B, C, or D). Do not include any explanation or additional text. \\
Answer:

\end{tcolorbox}
\caption{Prompt-Ask framework for downstream tasks - inducing curiosity}
\label{fig:promptask}
\end{figure}

\subsection{Qualitative Analysis of Follow-up Questions}
\label{sec:qual_followup}
We compare the NormAd questions (for prompt-ask vs adapter+prompt-ask models). Here are some high level takeways: 

In terms of question types and intents, Adapter follow-ups consist of a lot of yes/no norm checks: for example, ``is it rude to...?'', `` is it acceptable..?''. Prompt-based follow-ups are more explanatory probes -- ``How do norms in <country> treat...?'' This may explain the stronger performance of adapter-based models: they tend to ask more explicit and direct questions, improving contextual reasoning and cultural adaptability. Cultural grounding is more apparent in prompt-based followups, like ``In <country>, is \_\_\_ interpreted as \_\_ ? '' However, adapter follow-ups are more implicit. This maybe because these models are currently only trained in 3 countries, which are not present in the benchmarks we use for inference. In the future, we should focus on increasing the cultural grounding of fine-tuned models. Additionally, adapter models use more moralizing/judgment verbs (rude, bad manners, better host). While prompt-based methods use more interpretive verbs (suggest, indicate, align, conflict) that invite nuance. Finally, prompt-based methods ask more open-ended questions overall, however, they sometimes over-fit culture (presume norms, and use leading questions).

In the future, both methods can be improved: for adapter-based models, adding a light prompt scaffold that encourages why/how follow-ups and examples could enhance reasoning depth; for prompt-based models, introducing constraints to avoid presuppositions may lead to more precise and culturally grounded questions.

\section{Significance Tests Across Experiments}

\noindent \textbf{Linguistic Alignment.} Humans show larger dispersion across countries than LLMs (SD = 0.0785 vs. 0.029; reported F = 7.33), indicating significantly greater variance in human linguistic styles (very large N; variance ratio well above 1). 

\noindent \textbf{Topic preference Alignment.} \texttt{LLaMA-3-8B} shows the only positive average correlation with human topic rankings, other models are near-zero or negative. We find p = 0.043 significant; whereas negative values (e.g., GPT-4o $\rho = -0.29$) are also small in magnitude (|t| < 1.2).

\noindent \textbf{Inducing Curiosity (inherent curiosity)} Using 70 statements - adapter-FT asks questions on 75\% of prompts vs. 20\% for full-FT and 0\% for prompt-only; relevance is 98–100\% for all. Two-proportion tests with n = 70 per model show adapter vs. full-FT is highly significant (z = 6.5, $\rho$ < 10$^{-9}$), and adapter vs. prompt-only is modestly decisive.  

% \noindent \textbf{Downstream Tasks} For CulturalBench experiments across model types, we observe p = 2.3 x 10$^{-22}$. For NormAd, we have p = 2.38 x 10$^{-25}$ and for Cultural Commonsense, 

\section{Model Choices, Implementation Details and Computational Resources}

For our experiments, we selected a mix of open- and closed-source models to examine cultural curiosity across different architectures and training paradigms. Specifically, we used \texttt{LLaMA-3-8B}~\cite{dubey2024llama} for both prompt-based and fine-tuning experiments, focusing primarily on the smaller model due to its strong alignment with human data and lower computational overhead. We additionally evaluated \texttt{GPT-4o}~\cite{achiam2023gpt}, \texttt{GPT-5},\footnote{https://openai.com/index/introducing-gpt-5/} \texttt{LLaMA-3-70B}, \texttt{Claude-Sonnet-4},\footnote{https://www.anthropic.com/news/claude-4} and \texttt{Qwen-3-14B}~\cite{yang2025qwen3} to provide a broader comparison across model families. For fine-tuning, we applied LoRA adapters and full-fine-tuning to \texttt{LLaMA-3-8B} under different data configurations (e.g., country-specific vs. country-all, Yahoo vs. Reddit), using culture-aware and inherent curiosity objectives.

All experiments were run on NVIDIA A40 GPUs, using the Hugging Face Transformers and PEFT libraries. Fine-tuning followed standard hyperparameters, with a maximum sequence length of 1024 tokens and a fixed random seed for reproducibility. Adapter training was also chosen as they are more efficient, allowing rapid experimentation across multiple cultural configurations. In our experiments, we also find that adapter-based models do quite well in human-LLM curiosity alignment.

\section{Reproducibility}
We open-source our codes and data, which are uploaded to the submission system. This would help future work to reproduce our results and explore culture-aware curiosity in humans and LLMs, and its impact in downstream tasks.

\begin{table*}
\centering
\small
\caption{Dataset of Everyday Statements for Inherent Curiosity Evaluation}
\label{tab:statements}
\begin{tabular}{>{\raggedright\arraybackslash}p{0.15\textwidth} >{\raggedright\arraybackslash}p{0.75\textwidth}}
\toprule
\textbf{Category} & \textbf{Statements} \\
\midrule

Everyday Life & Alex arrived late to the meeting. \\
& Someone brought a gift to dinner without wrapping it. \\
& Sarah decided to sit at the back of the classroom. \\
& The teacher suddenly stopped talking in the middle of the lecture. \\
& Two friends started whispering during the movie. \\
\midrule
\addlinespace

Social  & Maria refused to shake hands at the event. \\
Situations & A guest walked straight into the kitchen at someone's house. \\
& John didn't respond when greeted by a colleague. \\
& A person started eating dessert before the main course. \\
& Someone left the party without saying goodbye. \\
\midrule
\addlinespace

Work \& School & The boss canceled the meeting at the last minute. \\
& Students stood up when the professor entered the room. \\
& A worker refused to answer emails after 5 p.m. \\
& A student called their teacher by a nickname. \\
& Employees were chatting loudly in the office. \\
\midrule
\addlinespace

Ethics \& & A child lied to their parents about homework. \\
Morality & A man kept extra change the cashier accidentally gave him. \\
& Someone ignored a homeless person asking for help. \\
& A person secretly recorded a private conversation. \\
& A passenger refused to give up a seat to an elderly person. \\
\midrule
\addlinespace

Science \& & Water can boil at room temperature. \\
Facts & Dinosaurs and humans lived at the same time. \\
& The sun could suddenly stop shining tomorrow. \\
& Some people believe time travel is possible. \\
& There are animals that can survive without oxygen. \\
\midrule
\addlinespace

Technology \&  & An AI refused to answer a user's question. \\
Future & A robot was programmed to laugh randomly. \\
& Someone trusted a self-driving car to take them home. \\
& A drone delivered a package late at night. \\
& People started wearing VR headsets in public. \\
\midrule
\addlinespace

Abstract /  & This statement is false. \\
Paradoxical & A ship has had every part replaced --- is it still the same ship? \\
& A person knows a secret that no one else can ever know. \\
& If a tree falls in a forest and no one hears it, does it make a sound? \\
& Someone dreamed they were awake. \\
\midrule
\addlinespace

Random /  & A stranger gave a child candy on the street. \\
Surprising & People clapped after a plane landed. \\
& A friend borrowed money and never returned it. \\
& Someone talked to their plants every morning. \\
& A cat suddenly barked like a dog. \\

\bottomrule
\end{tabular}
\end{table*}

\begin{table*}[htbp]
\centering
\small
\caption{Dataset of Cultural Statements (Country-wise) for Inherent Curiosity Evaluation}
\label{tab:statements_culture}
\begin{tabular}{>{\raggedright\arraybackslash}p{0.15\textwidth} >{\raggedright\arraybackslash}p{0.75\textwidth}}
\toprule
\textbf{Category} & \textbf{Statements} \\
\midrule

Brazil &  In Brazil, it is common for friends to greet each other with physical gestures. \\ &  
Brazilian families often spend weekends together in a particular way. \\ & 
Football matches in Brazil usually come with unique social rituals. \\ &
In Brazil, people tend to treat teachers in a distinctive manner. \\ &
A wedding in Brazil usually involves traditions beyond the ceremony. \\ & 
Brazilians often approach mealtimes differently from other cultures. \\ & 
A Brazilian student is expected to interact with professors in a certain way. \\ & 
During Carnival in Brazil, certain behaviors are widely accepted that might not be elsewhere. \\ & 
In Brazil, people have specific expectations about lateness. \\ &
Many Brazilian workplaces begin the day in a particular way.\\
\midrule
\addlinespace

Philippines &  In the Philippines, community celebrations often include unique practices. \\ &
Filipino families are expected to behave in certain ways during holidays. \\ &
Respect toward elders in the Philippines is shown through specific gestures. \\ &
Filipino classrooms often encourage behaviors different from Western norms. \\ &
Religious festivals in the Philippines are tied to specific community practices. \\ &
Neighbors in the Philippines usually interact in ways that reflect local values. \\ &
Filipino weddings often extend beyond the couple and include wider groups. \\ &
Hospitality in the Philippines takes forms not common elsewhere. \\ &
Gift-giving in the Philippines follows unspoken cultural rules. \\ &
Daily greetings in Filipino households involve rituals outsiders may overlook. \\
\midrule
\addlinespace

UK & In the UK, politeness takes a form that is not always obvious to outsiders. \\ & 
British families have unique expectations during Sunday gatherings. \\ & 
In UK workplaces, meetings often begin with a particular social convention. \\ & 
British weddings typically follow certain cultural patterns. \\ & 
Pubs in the UK play a social role that extends beyond drinking. \\ & 
In the UK, silence in conversations can mean something specific. \\ & 
British classrooms have norms that might surprise people from other cultures. \\ & 
Greetings in the UK are often understated in a characteristic way. \\ & 
British people have particular unwritten rules about queuing. \\ & 
Public discussions of politics in the UK follow a distinctive pattern. \\ 
% \midrule
% \addlinespace
\bottomrule
\end{tabular}
\end{table*}

\end{document}